\documentclass{article}

\usepackage{microtype}
\usepackage{graphicx}
\usepackage{subcaption}
\usepackage{booktabs} 
\usepackage{multirow}
\usepackage{multicol}

\usepackage{hyperref}

\usepackage[preprint]{icml2026}

\usepackage{amsmath}
\usepackage{amssymb}
\usepackage{mathtools}
\usepackage{amsthm}

\usepackage[capitalize,noabbrev]{cleveref}

\theoremstyle{plain}

\theoremstyle{definition}

\theoremstyle{remark}

\usepackage[textsize=tiny]{todonotes}

\icmltitlerunning{Submission and Formatting Instructions for ICML 2026}

\begin{document}

\twocolumn[
  \icmltitle{UniFluids: Unified Neural Operator Learning with Conditional Flow-matching}



  \icmlsetsymbol{equal}{*}

    \begin{icmlauthorlist}
    \icmlauthor{Haosen Li}{amss}
    \icmlauthor{Qi Meng}{amss}
    \icmlauthor{Jiahao Li}{msra}
    \icmlauthor{Rui Zhang}{ruc}
    \icmlauthor{Ruihua Song}{ruc}
    \icmlauthor{Liang Ma}{ioz}
    \icmlauthor{Zhi-Ming Ma}{amss}
    \end{icmlauthorlist}
    
    \icmlaffiliation{amss}{Academy of Mathematics and Systems Science, Chinese Academy of Sciences}
    \icmlaffiliation{msra}{Microsoft Research Asia}
    \icmlaffiliation{ruc}{Renmin University of China}
    \icmlaffiliation{ioz}{Institute of Zoology, Chinese Academy of Sciences}
    

    \icmlkeywords{Neural Operator, Conditional Flow Matching, PDE Forecasting, Foundation Model}





  \vskip 0.3in
]
\printAffiliationsAndNotice{}



\begin{abstract}
  Partial differential equation (PDE) simulation holds extensive significance in scientific research. Currently, the integration of deep neural networks to learn solution operators of PDEs has introduced great potential. In this paper, we present \textit{UniFluids}, a conditional flow-matching framework that harnesses the scalability of diffusion Transformer to unify learning of solution operators across diverse PDEs with varying dimensionality and physical variables. Unlike the autoregressive PDE foundation models, UniFluids adopts flow-matching to achieve parallel sequence generation, making it the first such approach for unified operator learning. Specifically, the introduction of a unified four-dimensional spatiotemporal representation for the heterogeneous PDE datasets enables joint training and conditional encoding. Furthermore, we find the effective dimension of the PDE dataset is much lower than its patch dimension. We thus employ $x$-prediction in the flow-matching operator learning, which is verified to significantly improve prediction accuracy. We conduct a large-scale evaluation of UniFluids on several PDE datasets covering spatial dimensions 1D, 2D and 3D. Experimental results show that UniFluids achieves strong prediction accuracy and demonstrates good scalability and cross-scenario generalization capability. The code will be released later.
\end{abstract}

\section{Introduction}
Partial differential equation (PDE) simulation holds extensive significance in scientific research. Currently, neural operators\cite{lu2021learning,li2020fourier,li2022fourier,han2018solving,li2023geometry,ye2024pdeformer,hao2023gnot,mccabe2023multiple,hao2024dpot,
chen2024omniarch}, which integrate deep neural networks to learn solution operators of PDEs, have shown strong performance on PDE benchmarks and shown great promise. The development of neural operators has evolved from initially learning a single type of parametric PDEs toward pre-training on heterogeneous datasets generated from various PDE families.
However, existing neural operators and PDE foundation models still face challenges in achieving accurate predictions and scalability when learning across diverse PDE families, variable sets, and spatial dimensions (1D,2D and 3D). First, most approaches rely on deterministic short-step autoregressive rollouts, where small errors compound over time~\cite{wang2025mixture}. Additionally, the training process based on point-to-point matching in original state space tends to average out variability, overlooking the uncertainty in practical PDEs and often causing over-smoothing and spectral (or energy) bias\cite{rahaman2019spectral,khodakarami2025mitigating}. Second, existing approaches either rely on dataset- or dimension-specific architectures and pipelines \cite{hao2024dpot,mccabe2023multiple,herde2024poseidon}, or represent the data in frequency domain with high-frequency truncation to support the heterogeneous datasets for training \cite{chen2024omniarch,zhang2025omnifluids}, which limits their generality or causes important high-frequency information loss.   

Recent advances in flow generative models - encompassing both diffusion models and flow-matching approaches - for spatiotemporal data modeling \cite{price2023gencast,davtyan2023efficient,ruhling2023dyffusion} offer an appealing alternative to deterministic operator rollouts. By modeling conditional distributions, these approaches enable parallel generation of a sequence of future states while maintaining high-fidelity sample quality and stable training dynamics. These strengths motivate us to adopt this paradigm for probabilistic operator learning. Yet, existing work on diffusion model-based neural operator \cite{hu2024wavelet,yang2023denoising} remain confined to specific datasets, motivating our pursuit of a unified and more generalizable framework.

In this work, we propose \emph{UniFluids}, a unified conditional generative operator for modeling complex spatiotemporal dynamics driven by multi-dimensional and multi-variable PDEs. UniFluids is based on conditional flow-matching approach together with Transformer architecture. In terms of the data representation, we introduce a unified spatiotemporal tensor interface that supports heterogeneous variable sets and a common formulation across various spatial dimensions (1D,2D and 3D), enabling joint training and across datasets without specializing the architecture. In terms of the condition encoding, a condition encoder together with temporal and spatial aggregation layers extracts both dense and compact aggregated conditioning from observed sequence of historical states, and a conditional flow-matching operator generates a sequence of future states in parallel, in order to alleviate autoregressive error cascades. Both the condition encoder and flow-matching operator are based on the Transformer architecture, ensuring global receptive fields across diverse PDE families and scalability. 

Notably, compared to other spatiotemporal data (e.g., the vision data \cite{ho2020denoising,karras2022elucidating,liu2022flow}), PDE data have higher dimensions and involve various multi-physics coupling, which increases the number of channels, the volume of patches and the heterogeneity in modeling.  However, we also have prior knowledge that PDE data patches exhibit strong correlations, indicating that they typically lie in low-dimensional subspace. The intuition is that simultaneously handling varying degrees of dimension degeneration across different datasets may affect the prediction accuracy. Our detailed analysis of effective dimensionality reveals that the discrepancy between effective dimensions and the original grid dimension grows substantially with increasing spatial dimension (from 1D to 3D). As the paper JiT ~\cite{li2025back} points out that this discrepancy worsens the optimization process and the prediction accuracy, we adopt the \emph{$x$-prediction} in the flow-matching operator learning: that is, parameterizing the original state $\mathbf{x}$ using a Transformer and defining the loss on velocity (i.e., the $\mathbf{v}$-loss which is calculated by the state $\mathbf{x}$) in flow-matching. Empirical investigation and ablation study show that the $\mathbf{x}$-prediction significantly improves both the accuracy and convergence.

The main {\bf contributions} are summarized as below:
\begin{itemize}
  \item \textbf{A unified conditional generative operator.}
  We present UniFluids, a conditional flow-matching operator with a unified spatiotemporal tensor representation that supports heterogeneous variable sets and spatial dimensions (1D,2D and 3D) within one model, enabling joint probabilistic modeling and parallel sequence generation across various PDE datasets.

 \item \textbf{A key finding on effective dimension and parameterization.} 
 Our approach mitigates the latent misalignment arising from varying degrees of dimension degeneration across the heterogeneous datasets through adopting the $\mathbf{x}$-prediction technique in training the conditional generative operator, which empirically shows significant error reduction in prediction. 

  \item \textbf{Large-scale evaluation.}
  We scale UniFluids up to 512M parameters and conduct experiments on several PDE benchmark datasets (mainly CFD related)  spanning 1D, 2D and 3D spatial dimensions. Experiments demonstrate that UniFluids reduces prediction error by up to 86.7\% compared to baselines and exhibits generalization capabilities, enabling accurate zero-shot inference on both in-domain and unseen out-of-domain downstream tasks.
\end{itemize}


\section{Related Work}

\subsection{Neural Operators for Parametric PDE learning}
Neural operators learn solution operators of parametric PDEs from data and have achieved strong results in applications such as
fluid dynamics and climate forecasting \citep{li2020fourier,pathak2022fourcastnet}.
To address diverse PDE structures and discretizations, many architectures have been proposed:
DeepONet factorizes operators into branch/trunk networks \citep{lu2021learning}, while works like FNO learn global couplings efficiently in the Fourier domain \citep{li2020fourier,wan2025spectral}.
Several extensions adapt operator learning to complex geometries or irregular meshes \citep{li2022fourier,liu2023nuno,li2023geometry}, and physics-informed variants incorporate PDE residual constraints to improve generalization \citep{raissi2019physics,li2024physics,wang2021learning,cuomo2022scientific,wang2024beno,fengheap}.
Transformer-based architecture for operator learning further introduces patchification and efficient attention mechanisms
\citep{cao2021choose,li2022transformer,hao2023gnot,guibas2021adaptive,dosovitskiy2020image,liu2021swin}.
Despite these advances, these approaches are typically limited to training on a specific PDE dataset with fixed dimensionality, and the time-dependent forecasting typically relies on deterministic short-step prediction via autoregressive rollout, which causes error accumulations and induces over-smoothing or spectral bias \citep{pathak2022fourcastnet,brandstetter2022message}.

\subsection{PDE Foundation Models}
Pre-training is a dominant paradigm in NLP and vision \citep{radford2019language,devlin2019bert,he2020momentum,he2022masked},
and is increasingly explored in scientific machine learning, such as weather forecasting and molecular modeling \citep{jumper2021highly,nguyen2023climax,zhou2023uni}.
For PDE data, early cross-system pretraining explores symmetry/contrastive objectives or more universal formulations \citep{mialon2023self,subramanian2023towards,yang2023context}.
Recent PDE foundation-model\cite{zhang2025omnifluids,mccabe2023multiple,hao2024dpot,chen2024omniarch,ye2024pdeformer} efforts scale pretraining across heterogeneous PDE datasets:
MPP proposes autoregressive pretraining across multiple physical systems \citep{mccabe2023multiple},
DPOT further scales with an autoregressive denoising objective and Fourier-enhanced transformer backbone \citep{hao2024dpot},
and OmniArch targets unified 1D/2D/3D modeling with a shared transformer core plus physics-alignment fine-tuning \citep{chen2024omniarch}.
These trends are enabled by standardized benchmarks and multi-domain datasets \citep{takamoto2022pdebench,gupta2022towards,luo2023cfdbench}.
However, most existing foundation models remain deterministic next-step (or short-window) predictors and rely on autoregressive generation at inference, making long-horizon forecasts vulnerable to compounding errors \citep{mccabe2023multiple,hao2024dpot,chen2024omniarch}.

\subsection{Flow-based Generative Models for Operator Learning}
Diffusion and score-based generative models \citep{sohl2015deep,ho2020denoising,song2020score,karras2022elucidating,jiang2025sig} learn complex distributions via stochastic processes and have been extended from images \cite{peebles2023scalable,li2025back} to videos\cite{ho2022video,he2022latent} and are also applied for neural operator learning \citep{lu2024generative,zhang2025physics,huang2024diffusionpde}.
Probability-flow ODE formulations connect these models to deterministic sampling with ODE solvers \citep{song2020score,chen2018neural},
while flow matching and rectified flows directly learn the probability-flow velocity field \citep{lipman2022flow,liu2022flow}.
These frameworks are adopted by neural operators \cite{price2023gencast,davtyan2023efficient,ruhling2023dyffusion} which enables parallel generation of a sequence of future states. However, most prior generative-model-based operator frameworks do not explicitly target unified modeling across heterogeneous PDE datasets, and typically require dataset-specific pre-processing, variable-specific encoding, or dimension-dependent designs.

\begin{figure*}[t]
  \centering
  \includegraphics[width=\textwidth]{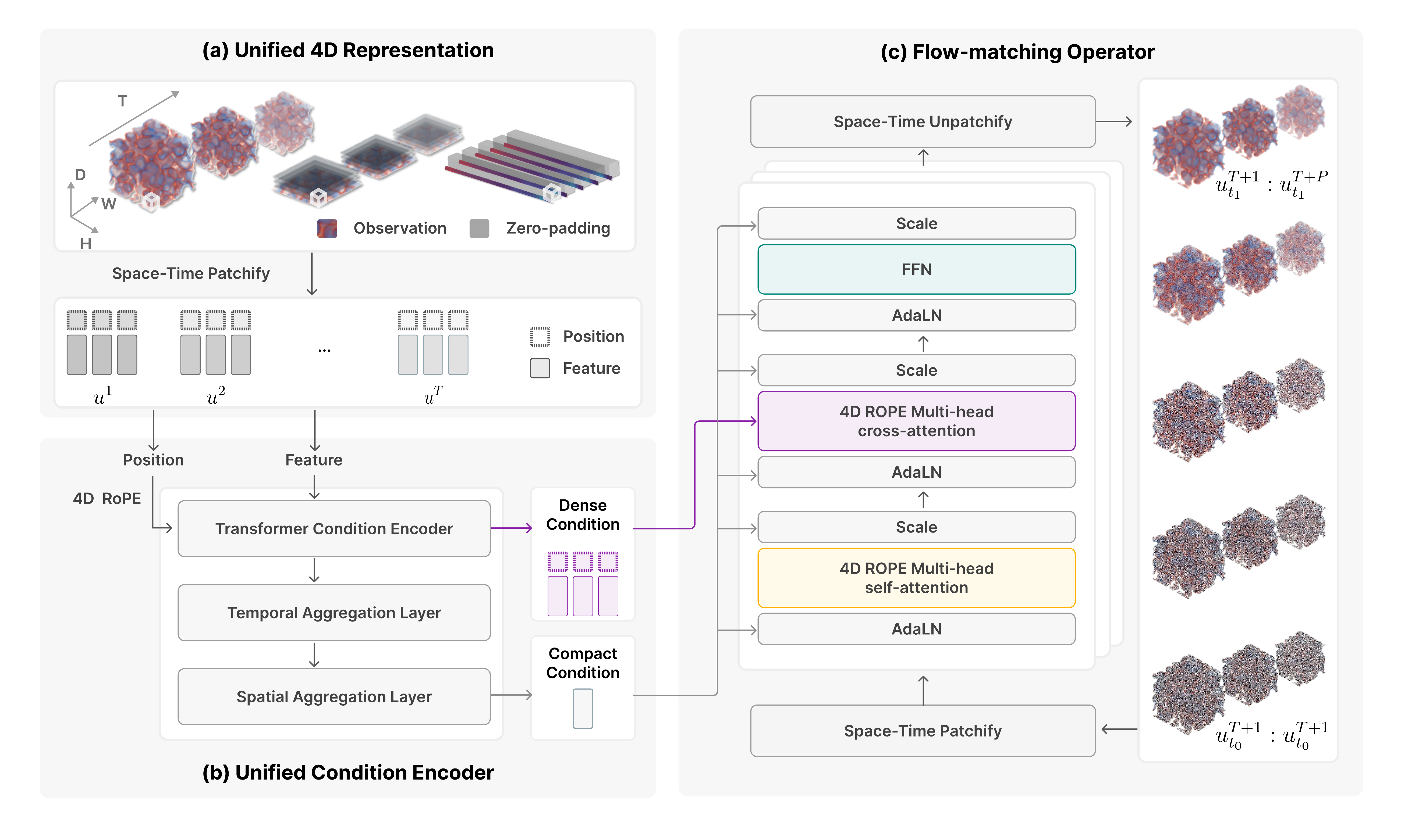}
  \vspace{-10mm}
  \caption{\textbf{Overall architecture.} 
    (a) \emph{Unified 4D representation:} heterogeneous 1D/2D/3D PDE trajectories with different variable sets are aligned to a canonical 4D grid \((t,h,w,d)\) with zero-padding on degenerate axes and a channel mask for missing variables, then space--time patchified into tokens paired with explicit 4D coordinates, enabling 4D RoPE. 
    (b) \emph{Unified condition encoder:} a Transformer encodes the observed history into \(\mathbf{c}_{\text{tok}}\) (dense condition) and a aggregated \(\mathbf{c}_g\) (compact condition) via temporal/spatial aggregation. 
    (c) \emph{Conditional flow-matching operator:} starting from noisy future-window tokens, the operator stacks 4D-RoPE self-attention and cross-attention to \(\mathbf{c}_{\text{tok}}\), with AdaLN modulated by \((t,\mathbf{c}_g,\texttt{dim\_type})\), to predict the clean solution.}
\vspace{-5mm}
  \label{fig:main-model}
\end{figure*}

\section{Methodology}
\subsection{Problem Setup}
\label{sec:problem_setup}

We consider time-dependent, parameterized PDE systems over a spatial domain $\Omega \subset \mathbb{R}^d$ ($d\in\{1,2,3\}$) and a time interval $\mathcal{T}=[0,T]$, with state field $u(x,t)\in\mathbb{R}^{m}$:
\begin{equation}
\label{eq:pde_general}
\begin{aligned}
\frac{\partial u}{\partial t}(x,t) - \mathcal{F}[u;\,\vartheta](x,t) &= 0, \quad (x,t)\in \Omega\times\mathcal{T},\\
u(x,0) &= u^{0}(x), \quad x\in\Omega,\\
\mathcal{B}[u](x,t) &= 0, \quad (x,t)\in \partial\Omega\times\mathcal{T},
\end{aligned}
\end{equation}
where $\mathcal{F}$ denotes a (generally nonlinear) differential operator that may depend on unknown parameters $\vartheta\in\Theta$ (e.g., coefficients, source terms, or equation types), $u^0(x)$ and $\mathcal{B}$ respectively denote the initial and boundary conditions. This formulation covers a broad range of PDE families, such as Burgers equation, shallow water equation, reaction--diffusion equation, and Navier--Stokes equation. The state field may contain multiple physical variables (abbreviated as \emph{multi-physics}), e.g., $(u,v)$ for reaction--diffusion, $(V_x,V_y,\rho,p)$ for 2D compressible Navier--Stokes.

\paragraph{Multiple PDE Datasets.}
We denote a collection of datasets drawn from simulating diverse PDE families as
$\mathcal{D}=\bigcup_{k=1}^{K}\mathcal{D}_k$, where each $\mathcal{D}_k=\{u_i\}_{i=1}^{N_k}$ contains $N_k$ trajectories (solution sequences) generated under a particular PDE and simulation configuration. A trajectory $u_i$ is a time series recorded over $S_i$ time steps: 
\begin{align}
\label{eq:traj_discrete}
u_i &\;=\; \big(u_i^{1},\ldots,u_i^{S_i}\big),
\end{align}where each $u_i^t, t\in[S_i]$ is a state field that is discretized on regular spatial grid $\mathcal{X}_i$:
\begin{align}
\label{eq:traj_discrete}
u_i^{t} & \;=\; \{u_i^t(x);  x \in \mathcal{X}_i\}, t\in[S_i].
\end{align}
In practical settings, one typically observes only state trajectories,  and the exact PDE forms and the parameters $\vartheta$ are not known. Therefore, we learn the temporal evolution dynamics from the trajectories for future state prediction. 

\paragraph{Conditional Generation of Future States.}
Due to the inherent uncertainty (e.g., caused by measurement error, noise or subgrid modeling) in predicting future states, we apply probabilistic generative models for future states generation, i.e., generating $P$ frames of future states through modeling the conditional distribution of future states for given $S$ frames of history states.  
Specifically, we denote the  history and future states as
\begin{equation*}
\label{eq:hist_future}
U_{\mathrm{hist}} := \big(u^{1},\cdots,u^T\big), \qquad
U_{\mathrm{fut}}  := \big(u^{T+1},\cdots,u^{T+P}\big).
\end{equation*}
Our goal is to learn a unified conditional generative operator that models the conditional distribution
\begin{equation}
\label{eq:cond_dist}
\mathbb{P}_\theta\!\left(U_{\mathrm{fut}} \mid U_{\mathrm{hist}}\right),
\end{equation}
shared across multi-physics and multi-dimensional PDE families. This setup subsumes deterministic forecasting as a special case, while enabling distributional prediction and parallel generation of a sequence of future states.

\subsection{Unified Spatiotemporal Representation}
\label{sec:unified-interface}
To enable joint training condition encoding across heterogeneous PDE datasets with varying spatial dimensions and physical variables, we embed every state trajectory into a \emph{unified 4D token space}, represented as a set of
feature--coordinate pairs \(\{(\mathbf{z}_i,\mathbf{p}_i)\}_{i=1}^{L}\).
We first cast the discretized state trajectory $u$ onto a canonical dense tensor
\(\mathbf{u}\in\mathbb{R}^{T\times H\times W\times D\times C_{\max}}\), where $C_{max}$ denotes the channel size of physical variables and the spatiotemporal coordinates \((t,h,w,d)\) are uniformly aligned as four axes. Compared to $d=3$, cases with $d=1,2$ have degenerate spatial axes, and that are zero-padded to the minimal extent required for patchification. Furthermore, we use
a discrete \texttt{dim\_type}\(\in\{1,2,3\}\) to record the original spatial dimensionality of $u$. To reconcile
heterogeneous variable sets, all datasets share the same channel index space of size \(C_{\max}\).
The missing variables are filled with zeros and indicated by a binary channel mask
\(\mathbf{m}\in\{0,1\}^{C_{\max}}\), which is used to block noise injection into nonexistent channels
and to compute masked losses over valid variables.

We then partition \(\mathbf{u}\) into non-overlapping spatiotemporal patches of size
\((p_t,p_h,p_w,p_d)\). The \(i\)-th patch block
\(\mathcal{P}_i(\mathbf{u})\in\mathbb{R}^{p_t\times p_h\times p_w\times p_d\times C_{\max}}\) is
vectorized and linearly projected into a token embedding \(\mathbf{z}_i\). Crucially, each token
retains an explicit 4D integer coordinate \(\mathbf{p}_i=[t_i,h_i,w_i,d_i]\) that indexes the patch
origin on the unified 4D grid and directly parameterizes \emph{4D rotary positional embeddings (4D
RoPE) \cite{su2024roformer}
} in attention. Under this representation, the trajectories with different spatial dimensions (1D, 2D and 3D) naturally occupy axis-aligned
subspaces of the same 4D grid (handled uniformly via \(\mathbf{p}_i\), \texttt{dim\_type}, and
\(\mathbf{m}\)). As a result, the model supports 
a unified spatiotemporal representation across dimensions, resolutions, and time horizons without dimension--specific positional encodings, enabling joint training and reuse of a single Transformer-based operator across heterogeneous PDE datasets. Compared to OmniArch \cite{chen2024omniarch} which represents 1D, 2D and 3D PDE datasets in the frequency domain with high-frequency truncation, our proposed representation preserves the original data information while adaptively compressing features during training, avoiding information loss.

\subsection{Unified Conditional Flow-Matching Operator}
\label{sec:method:operator}
Given $T$ frames of history states $U_{\mathrm{hist}}=\big(u^{1},\cdots,u^T\big)$ and $P$ frames of future states
$U_{\mathrm{fut}}=\big(u^{T+1},\cdots,u^{T+P}\big)$, we aim to learn a unified conditional generative operator that models
$\mathbb{P}_\theta\!\left(U_{\mathrm{fut}}\,\middle|\,U_{\mathrm{hist}}\right)$ under the unified spatiotemporal representation.
The conditional flow-matching operator in UniFluids consists of two modules: (i) a \emph{unified patch encoder} that extracts dense and compact conditioning representations from $U_{\mathrm{hist}}$,
and (ii) a \emph{coordinate-aware transformer operator} that performs conditional flow-matching on 4D patch tokens to generate $U_{\mathrm{fut}}$. Figure.~\ref{fig:main-model} illustrates the overall architecture of UniFluids.

\paragraph{Conditioning via a unified patch encoder.}
We extract conditioning information from the history states $U_{\mathrm{hist}}$ using a transformer-based patch encoder operating on the unified 4D patch tokenization introduced in Sec.~\ref{sec:unified-interface}. Specifically, we patchify $U_{\mathrm{hist}}$ into a sequence of spatiotemporal tokens together with their explicit 4D integer coordinates $\{\mathbf{p}_i=(t_i,h_i,w_i,d_i)\}$. The encoder applies several self-attention blocks over these tokens, where positional information is injected through \emph{4D RoPE} parameterized by $\mathbf{p}_i$. The output serves as \emph{dense} conditioning tokens $\mathbf{c}_{\text{tok}}$ for downstream cross-attention.

In addition to dense conditions, we compute a  \emph{compact} condition vector $\mathbf{c}_g$ via two lightweight aggregation layers. 
First, a \textbf{temporal aggregation} layer collapses the patch-time axis with Fourier-featured weighting, following the temporal pooling strategy in DPOT\cite{hao2024dpot}.
We denote the dense condition token at history timestep $t\in\{1,\dots,T\}$ as $\mathbf{c}_{\text{tok}}^{t,s}$ and $s$ denotes the patch index at time $t$. 
The temporal aggregation layer aggregates information for each patch as:
\begin{equation}
\label{eq:temp_agg}
\tilde{\mathbf{c}}_{s}
\;=\;
\sum_{t=1}^{T}
\mathbf{W}_{t}\,\mathbf{c}_{\text{tok}}^{t,s}\odot
\exp\!\left(-\,i\,\boldsymbol{\gamma}^{\top}\,\bar t\right),
\end{equation}
where $\mathbf{W}_{t}$ is a learnable linear map,
$\bar t=t/T$ is the normalized discrete timestep, and
$\exp(-i\,\boldsymbol{\gamma}^{\top}\bar t)$ implements a Fourier feature mapping of $\bar t$ .

Second, a \textbf{spatial aggregation} layer performs Perceiver-style pooling, i.e., a learnable query token attends to $\{\tilde{\mathbf{c}}_{s}\}$ to yield
\begin{equation}
\label{eq:spatial_agg}
\mathbf{c}_g \;=\; \mathrm{Attn}\!\left(\mathbf{q},\,\{\tilde{\mathbf{c}}_{s}\},\,\{\tilde{\mathbf{c}}_{s}\}\right).
\end{equation}
Overall, the encoder outputs $(\mathbf{c}_{\text{tok}},\mathbf{c}_g)$, enabling the generative operator to condition on both fine-grained dense context and a compact summary of the underlying dynamics.

\paragraph{Conditional flow-matching operator.}
Given conditioning representations $(\mathbf{c}_{\text{tok}},\mathbf{c}_g)$ extracted from $U_{\mathrm{hist}}$, we learn a conditional flow-matching operator for generating the future window $U_{\mathrm{fut}}$ in one shot.
Following the ODE view of flow-based generative modeling, we define a linear interpolation path between data and noise:
\begin{equation}
\label{eq:fm_path_xpred}
\mathbf{z}_t \;=\; t\,U_{\mathrm{fut}} \;+\; (1-t)\,\boldsymbol{\epsilon}, \qquad
t\in[0,1],\ \boldsymbol{\epsilon}\sim p_{\text{noise}}.
\end{equation}
We use Transformer-based model to predict the future field,
\begin{equation}
\label{eq:xpred}
\hat{U}_{\mathrm{fut}}
\;=\;
f_\theta(\mathbf{z}_t,t;\mathbf{c}_{\text{tok}},\mathbf{c}_g).
\end{equation}
As pointed out in \cite{li2025back}, predicting the future field $\hat{U}_{\mathrm{fut}}$ (named as \emph{$x$-prediction} in \cite{li2025back}) has benefit when the data are located on low-dimensional subspace. We defer the rationale of choosing $x$-prediction to Sec.~\ref{sec:method:param}.
For training, we supervise the model through the corresponding velocity induced by $\hat{U}_{\mathrm{fut}}$ under Eq.~\eqref{eq:fm_path_xpred},
\begin{equation}
\label{eq:v_from_xpred}
\hat{\mathbf{v}}_\theta(\mathbf{z}_t,t;\mathbf{c}_{\text{tok}},\mathbf{c}_g)
\;=\;
\frac{\hat{U}_{\mathrm{fut}}-\mathbf{z}_t}{(1-t)+\varepsilon},
\end{equation}
and minimize the standard flow-matching $\mathbf{v}$-loss
\begin{equation}
\label{eq:vloss}
\mathcal{L}_{v}
=\mathbb{E}_{t,\boldsymbol{\epsilon},U_{\mathrm{fut}}}\!
\left\|
\hat{\mathbf{v}}_\theta(\mathbf{z}_t,t;\mathbf{c}_{\text{tok}},\mathbf{c}_g)
-
\frac{U_{\mathrm{fut}}-\mathbf{z}_t}{(1-t)+\varepsilon}
\right\|_2^2.
\end{equation}
At inference, we solve the following ODE by using the probability-flow ODE sampler,
\begin{equation}
\label{eq:ode_sampling}
\frac{d\mathbf{z}_t}{dt}
=
\hat{\mathbf{v}}_\theta(\mathbf{z}_t,t;\mathbf{c}_{\text{tok}},\mathbf{c}_g),
\qquad \mathbf{z}_0\sim p_{\text{noise}},
\end{equation}
and take $\mathbf{z}_1$ as the generated $U_{\mathrm{fut}}$.

Architecturally, $f_\theta$ is implemented as a \emph{coordinate-aware transformer operator} on the unified 4D patch tokenization (Sec.~\ref{sec:unified-interface}).
We patchify $\mathbf{z}_t$ into 4D spatiotemporal tokens with explicit coordinates, embed $t$ and the \texttt{dim\_type} indicator, and apply a stack of transformer blocks.
Each block performs (i) 4D-RoPE self-attention over noisy tokens to model spatiotemporal interactions and (ii) cross-attention from noisy tokens to the conditioning tokens $\mathbf{c}_{\text{tok}}$, with conditioning injected through adaptive normalization (AdaLN) using $(t,\mathbf{c}_g)$.
A final projection layer followed by unpatchification maps the token outputs back to the 4D tensor representation, yielding $\hat{U}_{\mathrm{fut}}$ in Eq.~\eqref{eq:xpred}.

\subsection{The Effectiveness of $x$-Prediction}
\label{sec:method:param}
As introduced in previous section, the conditional flow-matching operator is trained with the standard flow-matching $\mathbf{v}$-loss (ref. Eqn.(\ref{eq:vloss})) but the neural network performs \emph{$x$-pred}: the network's output is to represent the state field $U_{\mathrm{fut}}$ and is analytically converted to the velocity $\mathbf{v}$ through Eqn.(\ref{eq:v_from_xpred}) and Eqn.(\ref{eq:ode_sampling}). The motivation for adopting \emph{$x$-pred} is based on observation from \cite{li2025back}: when the effective dimension of the data is low, parameterize $x$ using the neural network is more effective than parameter the velocity $v$ or the noise $\epsilon$ in flow matching. We next conduct analysis on the effective dimensions of the 1D, 2D and 3D data.

\begin{table}[h]
\centering
\footnotesize
\setlength{\tabcolsep}{5pt}
\renewcommand{\arraystretch}{1.05}
\caption{Effective-dimension diagnostics for PDE patch vectors. $V$ denotes the size of the patch vector.}
\vspace{-1mm}
\begin{tabular}{c r c r r r}
\toprule
Dim-$d$ & $V$ & Target & PR $\downarrow$ & EV90 $\downarrow$ & EV90/$d$ $\downarrow$ \\
\midrule
1D & 48   & $x$        & 1.14    & 1    & 0.0208 \\
1D & 48   & $\epsilon$ & 47.60   & 43   & 0.8958 \\
1D & 48   & $v$        & 1.75    & 19   & 0.3958 \\
\midrule
2D & 512  & $x$        & 1.29    & 2    & 0.0039 \\
2D & 512  & $\epsilon$ & 471.99  & 428  & 0.8359 \\
2D & 512  & $v$        & 18.05   & 276  & 0.5391 \\
\midrule
3D & 5120 & $x$        & 7.09    & 34   & 0.0066 \\
3D & 5120 & $\epsilon$ & 2763.10 & 2814 & 0.5496 \\
3D & 5120 & $v$        & 139.13  & 1928 & 0.3766 \\
\bottomrule
\end{tabular}
\vspace{2pt}

\label{tab:intrinsic_dim}
\vspace{-2mm}
\end{table}

\begin{figure}[h!]
\vspace{-1mm}
\centering
\includegraphics[width=0.4\textwidth]{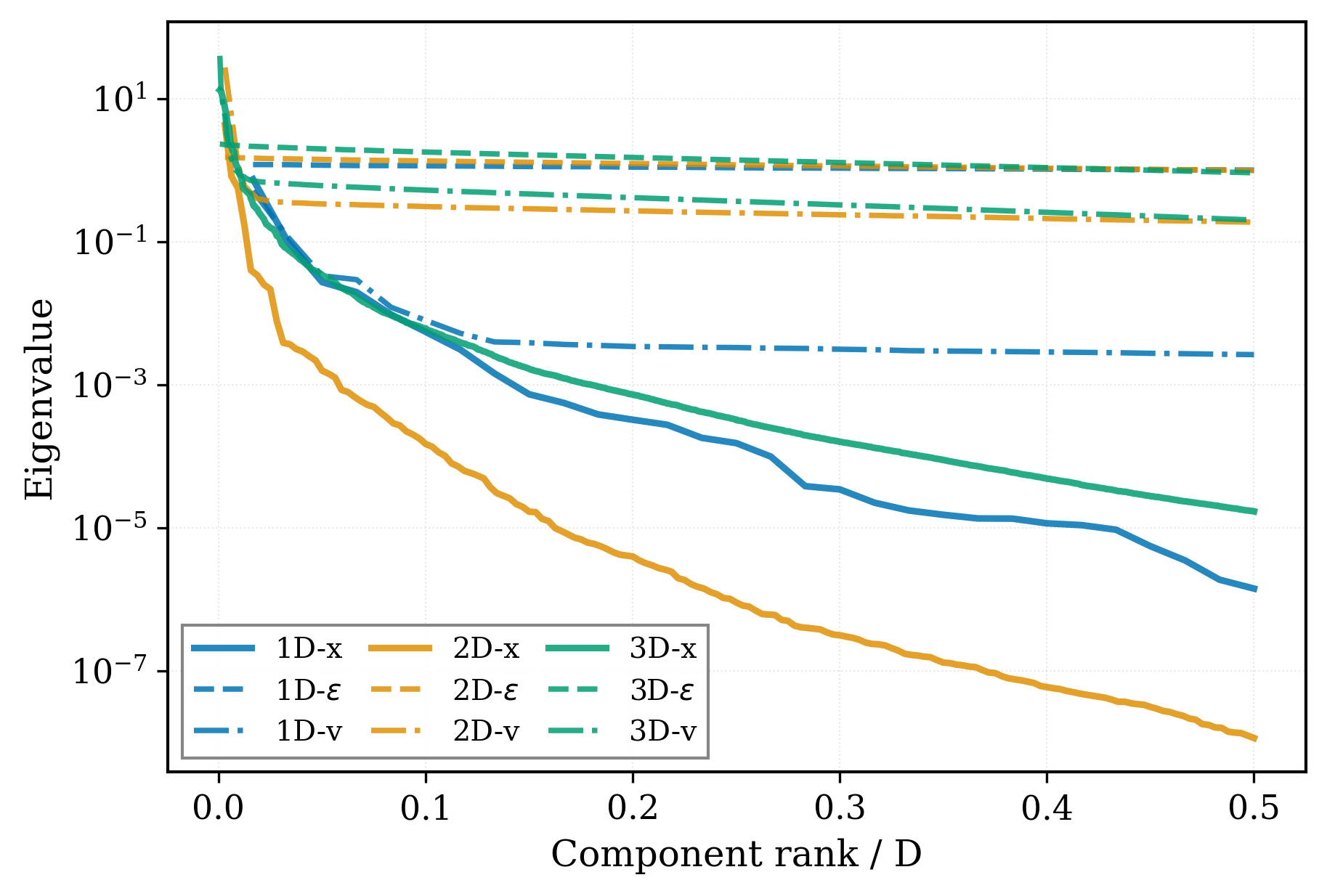}
\vspace{-1mm}
\caption{\textbf{Eigen-spectra of covariance for PDE patch vectors} (log-scale), comparing targets $x$, $\epsilon$, and $v$ across 1D/2D/3D.
Patchified PDE states concentrate variance in a small number of components (fast spectral decay),
whereas $\epsilon$ (and partially $v$) spans a much larger effective dimension, making $\epsilon$/$v$ prediction increasingly ill-conditioned as patch dimension grows.}
\label{fig:intrinsic_spectrum}
\vspace{-4mm}
\end{figure}

\paragraph{Diagnostics on PDE patches.}
We randomly sample $n{=}6000$ patch vectors from trajectories of the CFD-1D/2D/3D subsets (compressible Navier--Stokes equation) of PDEBench. We apply per-vector root-mean-square (RMS) normalization, and compute the covariance eigen-spectrum $\{\lambda_i\}_{i=1}^V$, where $V$ denotes the size of the patch vector.
We summarize effective dimension using the participation ratio $\mathrm{PR}=\frac{(\sum_i \lambda_i)^2}{\sum_i \lambda_i^2}$, $\mathrm{EV90}$ (components explaining 90\% variance) and $\frac{\mathrm{EV90}}{d}$.
Table~\ref{tab:intrinsic_dim} and Fig.~\ref{fig:intrinsic_spectrum} show that $\mathbf{x}$ (i.e., the patch vectors) has rapid spectral decay, indicating very low effective dimension, while $\boldsymbol{\epsilon}$ is close to full-dimensional and $\mathbf{v}$ lies in between. The disparity between the effective dimension and patch size becomes more pronounced as spatial dimension grows from 1D to 2D/3D.

\paragraph{Implication.}
Several reasons can cause the low effective dimension, such as correlations across spatial-temporal axes and inter-channel dependencies. Under the large patch size (e.g., $5120$ for 3D case), directly predicting $\boldsymbol{\epsilon}$ or $\mathbf{v}$ requires modeling high-dimensional and near-isotropic targets, which is empirically unstable at fixed model width. The diagnostic results on effective dimension of the patch vectors motivate us to adopt the \emph{x-pred}. By predicting $\mathbf{x}$, the network focuses on the structured low-dimensional component and filters isotropic noise.
This \emph{does not} change generation: we still obtain $\mathbf{v}$ via analytic conversion and integrate the same ODE. The ablation study in Sec.~\ref{sec:ablation_output_param} shows the effectiveness of the \emph{x-pred}.



\begin{table*}[t]
  \caption{Comparison of the relative $L^2$-error on the PDEBench subsets used in this work (lower is better). ``--'' indicates that the result is not reported by the corresponding baseline under the same setting. For the 2D CFD datasets, the parameters $(\eta,\zeta)=(0.1,0.1)$.
  The improvement is calculated by the best results of UniFluids and best results of the unified pre-training baselines.}
  \label{tab:unifluids_pdebench_main}
  \begin{center}
    \begin{small}
      \begin{sc}
      \resizebox{\textwidth}{!}{
        \begin{tabular}{lccccccccc}
          \toprule
          \multirow{2}{*}{Methods}
          & \multicolumn{3}{c}{1D}
          & \multicolumn{3}{c}{2D CFD}
          & \multicolumn{2}{c}{2D}
          & \multicolumn{1}{c}{3D} \\
          \cmidrule(lr){2-4}\cmidrule(lr){5-7}\cmidrule(lr){8-9}\cmidrule(lr){10-10}
          & CFD & Adv. & Bur.
          & $M=1$ & $M=0.1$ & All
          & SWE & Incom.
          & CFD (TURB.) \\
          \midrule

          \multicolumn{10}{l}{\textit{Baselines -- Task-specific  models}} \\
          \midrule
          \textbf{PINNs}
          & -- & 0.8130 & 0.9450
          & -- & -- & --
          & 0.0170 & --
          & -- \\
          \textbf{U-Net}
          & 2.6700 & 0.7760 & 0.3201
          & -- & -- & --
          & 0.0830 & 1.1200
          & 0.7989 \\
          \textbf{FNO}
          & 1.4100 & 0.0091 & 0.0174
          & -- & -- & --
          & 0.0044 & 0.2574
          & \textbf{0.3052} \\
          \midrule

          \multicolumn{10}{l}{\textit{Baselines -- Unified pre-training}} \\
          \midrule
          \textbf{MPP-L}
          & -- & -- & --
          & -- & -- & --
          & 0.0022 & --
          & -- \\
          \textbf{DPOT-L}
          & -- & -- & --
          & 0.0100 & 0.0087 & 0.0094
          & 0.0020 & --
          & -- \\
          \textbf{OmniArch-B}
          & 0.0340 & 0.0238 & \underline{0.0089}
          & -- & -- & --
          & \underline{0.0016} & 0.1726
          & 0.5209 \\
          \textbf{OmniArch-L}
          & \underline{0.0250} & 0.0182 & \textbf{0.0063}
          & -- & -- & --
          & \textbf{0.0014} & 0.1494
          & 0.4531 \\
          \midrule

          \multicolumn{10}{l}{\textit{Unified pre-training on 1D/2D/3D data (ours)}} \\
          \midrule
          \textbf{UniFluids-S}
          & 0.0270 & 0.0050 & 0.0152
          & 0.0086 & 0.0071 & 0.0079
          & 0.0037 & 0.0303
          & 0.4509 \\

          \textbf{UniFluids-M}
          & 0.0274 & 0.0043 & 0.0131
          & 0.0063 & 0.0050 & 0.0057
          & 0.0033 & 0.0229
          & 0.4483 \\

          \textbf{UniFluids-L}
          & 0.0264 & \underline{0.0037} & 0.0101
          & \underline{0.0052} & \underline{0.0038} & \underline{0.0045}
          & 0.0031 & \underline{0.0224}
          & 0.4565 \\

          \textbf{UniFluids-XL}
          & \textbf{0.0247} & \textbf{0.0028} & 0.0104
          & \textbf{0.0046} & \textbf{0.0033} & \textbf{0.0040}
          & 0.0024 & \textbf{0.0198}
          & \underline{0.4155} \\
          \midrule
          \textbf{Improvement}
          & \textbf{1.2\%} & \textbf{+69.2\%} & \textcolor{gray}{\textbf{-60.3\%}}
          & \textbf{+54.0\%} & \textbf{+62.1\%} & \textbf{+57.4\%}
          & \textcolor{gray}{-71.4\%} & \textbf{+86.7\%} & \textbf{+9.1\%}  \\
          \bottomrule
        \end{tabular}
      }
      \end{sc}
    \end{small}
  \end{center}
  \vskip -0.1in
\end{table*}

\section{Experiments}
\subsection{Experimental Setup}
\paragraph{Datasets.}
We primarily evaluate the performance of UniFluids on PDEBench~\cite{takamoto2022pdebench}, which covers  1D, 2D and 3D spatiotemporal PDE families with heterogeneous variable sets.
Our evaluation suite includes:
(i) \emph{1D Advection equation} and \emph{1D Burgers equation}, both modeling the evolution of a 1D scalar velocity field $V_x$;
(ii) \emph{1D compressible Navier--Stokes equation} (CFD-1D) with variables $(V_x, \rho, p)$, where $\rho$ denotes density and $p$ denotes pressure;
(iii) \emph{2D compressible Navier--Stokes equation} (CFD-2D) spanning multiple physical regimes, with variables $(V_x, V_y, \rho, p)$;
(iv) \emph{2D incompressible Navier--Stokes equation} (Incom-2D) with $(V_x, V_y)$ and a particle tracer field (\texttt{particles});
(v) \emph{Shallow-water equation} (SWE) with water depth $h$ (\texttt{water\_depth});
(vi) \emph{Reaction--diffusion equation} with activator $u$ and inhibitor $v$;
and (vii) \emph{3D compressible Navier--Stokes equation} (CFD-3D) with $(V_x, V_y, V_z, \rho, p)$.


\paragraph{Baselines.}
We compare against two categories of baselines: (i)\emph{Task-specific models.}
These methods are trained \emph{from scratch} for each individual dataset listed above and thus do not share knowledge across tasks.
Following the work \cite{chen2024omniarch}, we include representative physics-informed and operator-learning baselines: PINNs~\cite{raissi2019physics}, U-Net~\cite{ronneberger2015u}, and FNO~\cite{li2020fourier}.
(ii) \emph{Unified pre-training models.}
These methods perform unified learning on multi-physics corpora to acquire transferable representations, and then adapt to downstream cases.
We include Multiple Physics Pretraining (MPP)~\cite{mccabe2023multiple}, DPOT~\cite{hao2024dpot}, and OmniArch~\cite{chen2024omniarch} as representative unified modeling/pre-training baselines.

\paragraph{Training details.}
Unless otherwise specified, we use a single set of optimization hyperparameters across all datasets.
We train the model end-to-end with AdamW optimizer and a linear warmup followed by cosine learning-rate decay.
We apply mixed-precision training and gradient clipping for stability.
Under the conditional flow-matching objective, we sample the continuous noise level using a logit-normal parameterization, and employ condition dropout\cite{ho2022classifier} to enable classifier-free guidance at inference time.
For generation, we solve the corresponding probability flow ODE using a fixed-step Euler solver with a fixed number of steps.

\subsection{Main results}
We evaluate UniFluids on the PDEBench subsets spanning 1D/2D/3D dynamics under a unified training recipe, and report the relative $L^2$-error (lower is better). Table~\ref{tab:unifluids_pdebench_main} summarizes the results, comparing against (i) task-specific models trained per dataset, and (ii) unified pre-training baselines.

Overall, UniFluids achieves the strongest performance on most regimes with unified spatiotemporal modeling. On 1D Advection, UniFluids-XL reaches $0.0028$, yielding a $69.2\%$ error reduction over the best reported baseline under the same setting. For 2D compressible CFD, UniFluids-XL consistently improves upon DPOT-L across all regimes: it achieves up to $62.1\%$ error reduction compared to DPOT-L. On 2D incompressible Navier--Stokes, UniFluids-XL significantly outperforms unified pre-training baselines, improving over OmniArch-L from $0.1494$ to $0.0198$, corresponding to $86.7\%$ error reduction.
On the subsets 1D CFD and 3D CFD, UniFluids remains competitive despite using a single unified architecture and training objective. In particular, UniFluids achieves near-best performance on 1D CFD and SWE, while providing a strong unified solution that avoids dataset-specific designs. In 3D CFD, UniFluids-XL improves over OmniArch variants, indicating the benefit of unified conditional generative operator learning and \emph{$x$-pred} in high-dimensional regimes.

We visualize the predicted results of UniFluids on all data subsets and show them in Figure~\ref{fig:more_qual_1d}-\ref{fig:more_qual_3d} in Appendix. From the visualization, we can conclude that UniFluids accurately models the spatiotemporal dynamics of the PDEs.

\subsection{Scaling experiments}
\begin{figure}[t]
  \centering
  \includegraphics[width=0.4\textwidth]{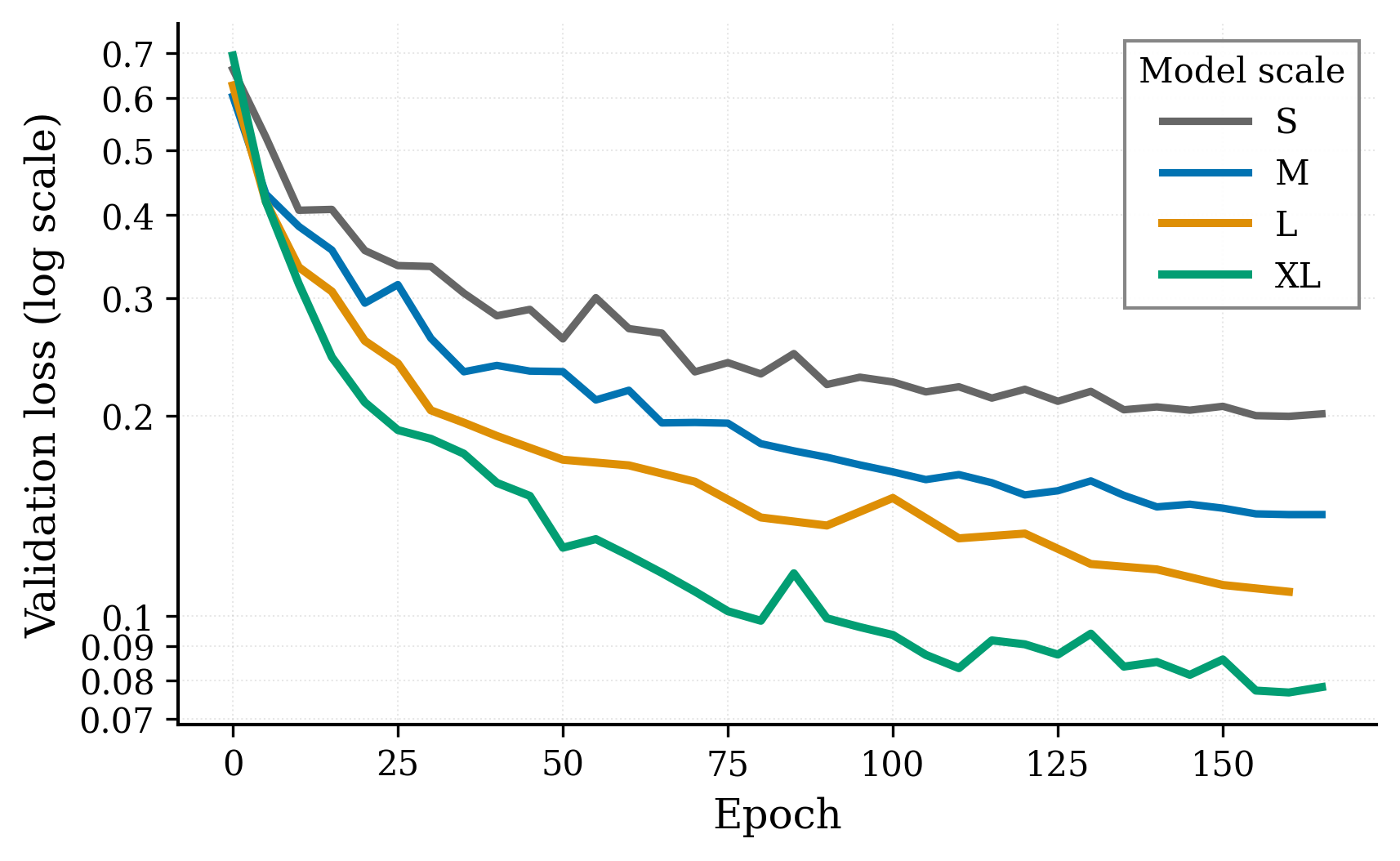}
    \vspace{-3mm}
  \caption{\textbf{Scaling behavior of validation loss.} Validation loss (log scale) across training epochs for UniFluids models of increasing scale (S/M/L/XL). Larger models consistently achieve lower validation loss, and the log-scale axis highlights the widening performance gap as training progresses.}
  \label{fig:val_loss_scales}
  \vspace{-4mm}
\end{figure}

To study scalability, we train four model sizes (S/M/L/XL) (see details in Table~\ref{tab:model_sizes} in Appendix) with the same training recipe, scaling capacity by increasing the embedding width and the depth of both the condition encoder and the flow-matching operator. All variants share the same 4D patch size $(p_t,p_h,p_w,p_d)=(2,8,8,8)$ and MLP ratio of $4.0$, with the number of attention heads following model width. Figure~\ref{fig:val_loss_scales} shows the validation loss over all PDE types. We observe that larger models consistently achieve lower validation loss (log-scale) and the gap widens over training, indicating favorable scaling behavior. This trend transfers to downstream accuracy (Table~\ref{tab:unifluids_pdebench_main}): UniFluids-XL improves 2D CFD \emph{All} from $0.0079$ to $0.0040$ and 2D incompressible Navier--Stokes from $0.0303$ to $0.0198$, suggesting that increased capacity is effectively utilized under the unified representation across heterogeneous PDE families.

\begin{figure*}[t]
  \centering
  \begin{minipage}[t]{0.3\textwidth}
    \centering
    \includegraphics[width=\linewidth]{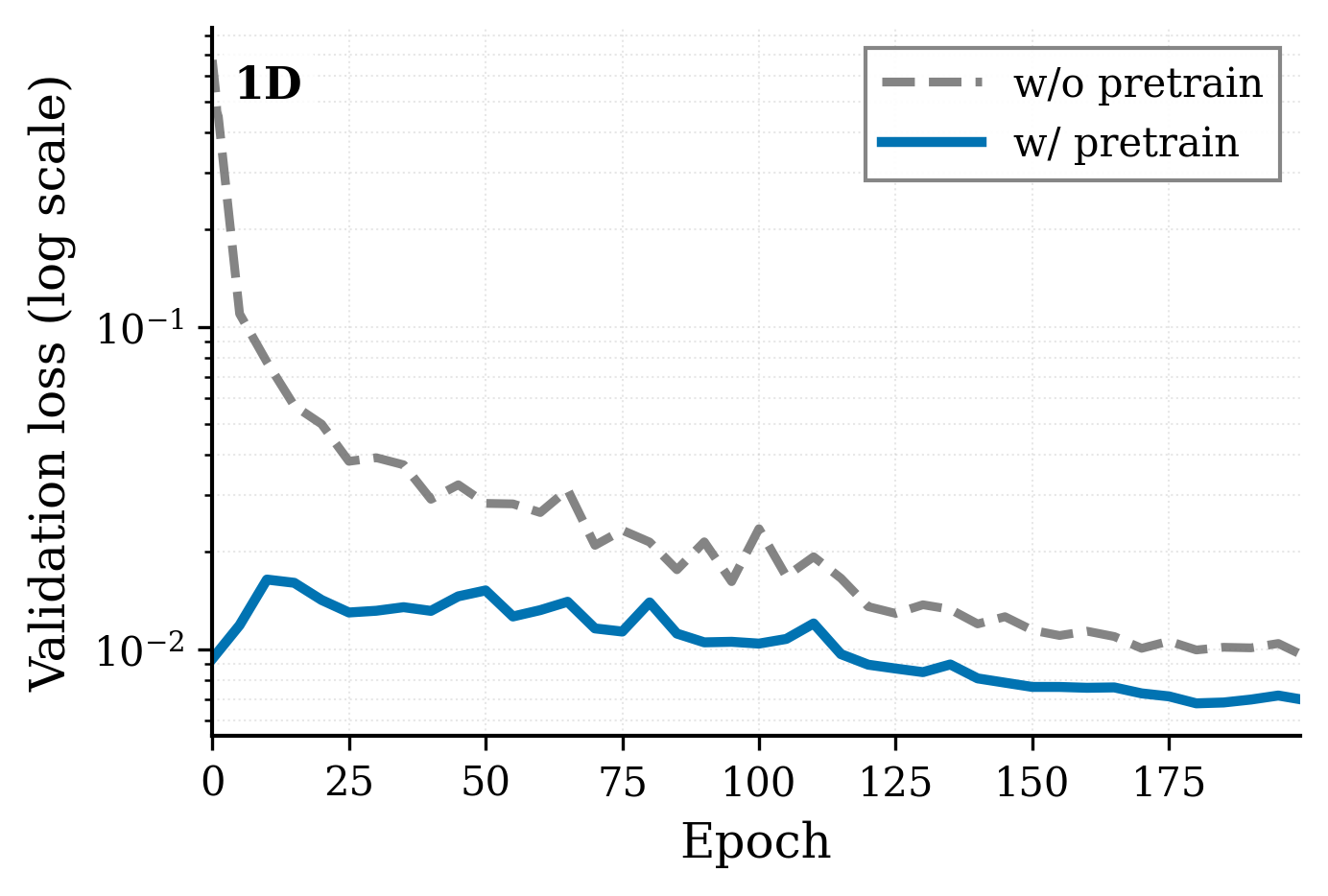}
    \vspace{-6mm}
    \caption*{\small 1D CFD}
  \end{minipage}
  \hfill
  \begin{minipage}[t]{0.3\textwidth}
    \centering
    \includegraphics[width=\linewidth]{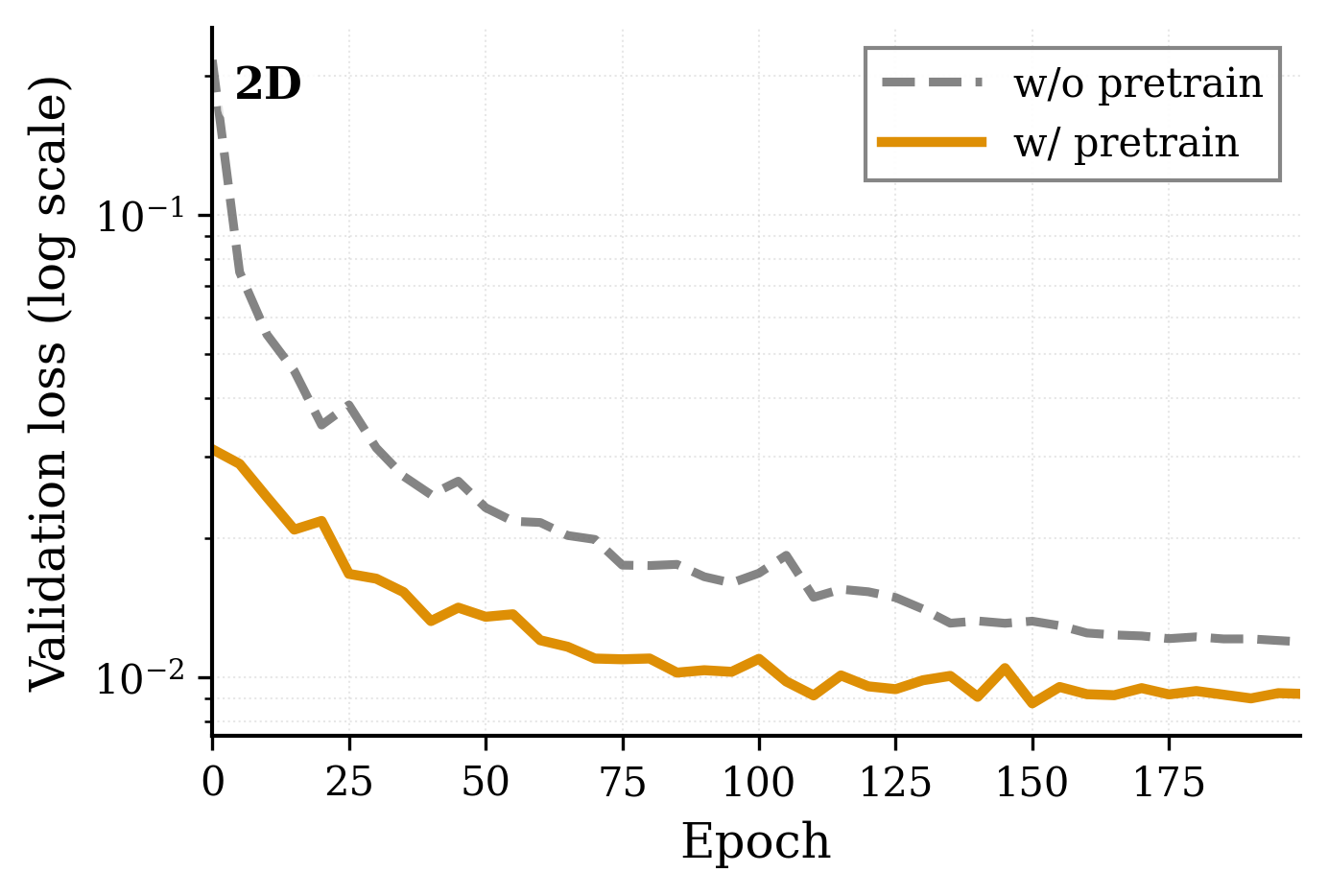}
    \vspace{-6mm}
    \caption*{\small 2D CFD}
  \end{minipage}
  \hfill
  \begin{minipage}[t]{0.3\textwidth}
    \centering
    \includegraphics[width=\linewidth]{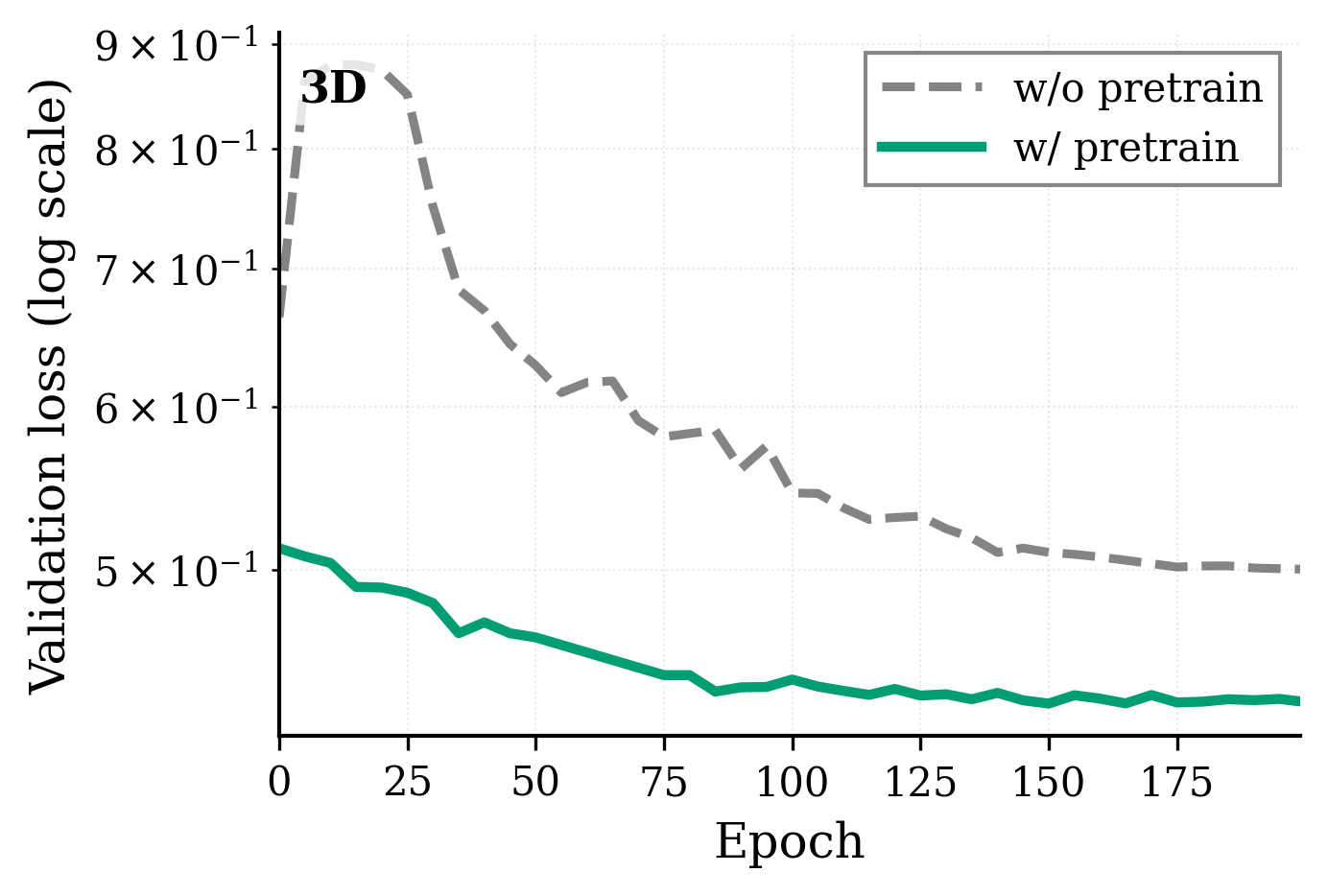}
    \vspace{-6mm}
    \caption*{\small 3D CFD}
  \end{minipage}
  \vspace{-1mm}
  \caption{Validation loss versus epoch on CFD fine-tuning (S model), comparing training from scratch (w/o pretrain) and fine-tuning from the foundation checkpoint (w/ pretrain) across 1D/2D/3D.}
  \label{fig:ft_val_loss_1d2d3d}
  \vspace{-2mm}
\end{figure*}

\begin{table*}[h!]
  \caption{Prediction-target ablation (S model). nRMSE $\downarrow$ on representative PDEBench subsets; ``2D CFD (All)'' averages the four CFD regimes in our suite. $\epsilon$-pred is omitted because training becomes numerically unstable (loss blows up to NaN/Inf under our unified pipeline), so no valid nRMSE can be obtained.}
  \label{tab:ablation_pred_target}
  \centering
  \footnotesize
  \setlength{\tabcolsep}{4pt}
  \renewcommand{\arraystretch}{1.08}
  \begin{sc}
  \begin{tabular}{@{}lccccccc@{}}
    \toprule
    \multirow{2}{*}{Method} &
    \multicolumn{3}{c}{1D} &
    \multicolumn{3}{c}{2D} &
    \multicolumn{1}{c}{3D} \\
    \cmidrule(lr){2-4}\cmidrule(lr){5-7}\cmidrule(lr){8-8}
    & Adv. & Bur. & CFD
    & CFD (All) & SWE . & Incom.
    & CFD (turb.) \\
    \midrule
    \emph{x-pred} &
    \textbf{0.0052} & 0.0253 & \textbf{0.0267} &
    \textbf{0.0561} & \textbf{0.0099} & \textbf{0.0553} &
    0.6749 \\
    \emph{v-pred} &
    0.0706 & \textbf{0.0243} & 0.0635 &
    0.8197 & 0.0334 & 0.5258 &
    \textbf{0.5545} \\
    \bottomrule
  \end{tabular}
  \end{sc}
\end{table*}

\subsection{Transferability and Generalization of UniFluids}
\label{sec:transfer_generalization}

We evaluate the benefit of pretraining UniFluids from two aspects:
(i) \emph{in-domain} adaptation efficiency on CFD, and (ii) \emph{out-of-domain} zero-shot generalization to unseen cases without additional training.
Note that the target CFD datasets are included during pretraining; thus the in-domain experiment isolates the effect of pretraining as an initialization under the same data distribution.
\paragraph{In-domain adaptation on CFD.}
We study whether a unified pretrained operator can be efficiently specialized to a single domain.
Specifically, we fine-tune UniFluids-S on 1D/2D/3D CFD and compare against training from scratch under identical architectures and optimization settings.
As shown in Figure~\ref{fig:ft_val_loss_1d2d3d}, fine-tuning starts from a substantially lower loss and converges faster, indicating that pretraining provides a strong initialization for learning the CFD dynamics.
Table~\ref{tab:ft_results} further reports the final evaluation in terms of nRMSE, demonstrating that fine-tuning consistently outperforms the training from scratch across CFD benchmarks.




\begin{table}[t]
  \caption{In-domain CFD adaptation (S model). nRMSE $\downarrow$. The 2D result is reported on the subset $(M,\eta,\zeta)=(1,10^{-1},10^{-1})$.}
  \label{tab:ft_results}
  \centering
  \footnotesize
  \setlength{\tabcolsep}{4pt}
  \renewcommand{\arraystretch}{1.08}
  \begin{sc}
  \begin{tabular}{@{}lccc@{}}
    \toprule
    Method & 1D CFD & 2D CFD & 3D CFD \\
    \midrule
    w/o pretrain & 0.0269 & 0.0063 & 0.5954 \\
    w/  pretrain & \textbf{0.0245} & \textbf{0.0049} & \textbf{0.5456} \\
    \midrule
    Reduction (\%) & 8.9\% & 22.2\% & 8.4\% \\
    \bottomrule
  \end{tabular}
  \end{sc}
  \vspace{-2mm}
\end{table}
\begin{table}[h!]
  \caption{Zero-shot performance on unseen cases (without fine-tuning). nRMSE $\downarrow$ (lower is better).}
  \label{tab:zeroshot}
  \centering
  \footnotesize
  \setlength{\tabcolsep}{4pt}
  \renewcommand{\arraystretch}{1.0}
  \begin{sc}
  \resizebox{0.85\linewidth}{!}{
  \begin{tabular}{@{}lccc@{}}
    \toprule
    Methods & 2D-Shock & 2D-KH & 2D-TOV \\
    \midrule
    FNO   & 0.7484 & 1.0891 & 0.5946 \\
    U-Net & 1.6667 & 0.1677 & 0.4217 \\
    MPP-L & 0.3243 & 1.3261 & 0.3025 \\
    \midrule
    \textbf{UniFluids-XL} & \textbf{0.2160} & \textbf{0.3113} & \textbf{0.2159} \\
    \bottomrule
  \end{tabular}
  }
  \end{sc}
  \vspace{-4mm}
\end{table}

\paragraph{Zero-shot generalization to unseen cases.}
We further evaluate the UniFluids in a zero-shot setting by directly applying the pretrained model to unseen cases without any additional training.
Table~\ref{tab:zeroshot} reports the nRMSE on three representative 2D unseen benchmarks: Shock wave dynamics, Kelvin--Helmholtz (KH) equation, and Orszag--Tang vortex (TOV).
Overall, UniFluids-XL achieves strong zero-shot performance and remains competitive against specialized baselines, demonstrating promising out-of-distribution zero-shot generalization.

\subsection{Ablation on $x$-Prediction}
\label{sec:ablation_output_param}

We ablate the output parameterization of the denoiser in conditional flow-matching using \texttt{UniFluids-S}.
We compare predicting the clean field $\mathbf{x}$ (\emph{x-pred}), the velocity field $\mathbf{v}$ (\emph{v-pred}), and the noise $\epsilon$ (\emph{$\epsilon$-pred}).
For fair comparison, \emph{all} variants use the loss computed on the velocity $\mathbf{v}$ even when predicting $x$ or $\epsilon$. We convert the output of the network into $\mathbf{v}$ and perform ODE-based sampling.
Table~\ref{tab:ablation_pred_target} reports nRMSE on representative PDEBench subsets.
Overall, \emph{x-pred} is consistently stable and accurate across all the datasets, while \emph{v-pred} is markedly less robust (especially on 2D CFD), and \emph{$\epsilon$-pred} fails to converge.

\section{Conclusion}
We introduced \textit{UniFluids}, a conditional flow-matching operator designed for unified modeling of solution operators across the heterogeneous PDE datasets. It enables the generation of PDEs solutions with multi-dimensional and multi-physics through a shared model. The adopted \emph{x-pred} in flow-matching to handle the low effective dimensional PDE data significantly improves the prediction accuracy. Furthermore, the model displays considerable scalability and out-of-domain generalization capabilities. Several directions merit future exploration, such as incorporating additional knowledge in the training process and leveraging single-step generation techniques to accelerate sampling.


\section*{Impact Statement}
This paper aims to advance machine learning methods for learning probabilistic solution operators of partial differential equations.
If successful, the proposed approach could reduce the computational cost of PDE simulation and enable faster scientific and engineering workflows (e.g., fluid dynamics and related physical modeling).
Potential risks include misuse of model predictions as ground truth in safety-critical settings, failure under distribution shift or long-horizon extrapolation, and dual-use concerns in applications such as defense-related simulation.
We encourage responsible use, including uncertainty-aware evaluation, out-of-distribution testing, and domain-expert validation when deployed in high-stakes scenarios.






\bibliography{example_paper}

@article{li2020fourier,
  title={Fourier neural operator for parametric partial differential equations},
  author={Li, Zongyi and Kovachki, Nikola and Azizzadenesheli, Kamyar and Liu, Burigede and Bhattacharya, Kaushik and Stuart, Andrew and Anandkumar, Anima},
  journal={arXiv preprint arXiv:2010.08895},
  year={2020}
}

@article{pathak2022fourcastnet,
  title={Fourcastnet: A global data-driven high-resolution weather model using adaptive fourier neural operators},
  author={Pathak, Jaideep and Subramanian, Shashank and Harrington, Peter and Raja, Sanjeev and Chattopadhyay, Ashesh and Mardani, Morteza and Kurth, Thorsten and Hall, David and Li, Zongyi and Azizzadenesheli, Kamyar and others},
  journal={arXiv preprint arXiv:2202.11214},
  year={2022}
}

@article{lu2021learning,
  title={Learning nonlinear operators via DeepONet based on the universal approximation theorem of operators},
  author={Lu, Lu and Jin, Pengzhan and Pang, Guofei and Zhang, Zhongqiang and Karniadakis, George Em},
  journal={Nature machine intelligence},
  volume={3},
  number={3},
  pages={218--229},
  year={2021},
  publisher={Nature Publishing Group UK London}
}

@article{li2025back,
  title={Back to basics: Let denoising generative models denoise},
  author={Li, Tianhong and He, Kaiming},
  journal={arXiv preprint arXiv:2511.13720},
  year={2025}
}

@article{wang2025mixture,
  title={Mixture-of-Experts Operator Transformer for Large-Scale PDE Pre-Training},
  author={Wang, Hong and Xin, Haiyang and Wang, Jie and Yang, Xuanze and Zha, Fei and Dong, Huanshuo and Jiang, Yan},
  journal={arXiv preprint arXiv:2510.25803},
  year={2025}
}

@article{li2022fourier,
  title={Fourier neural operator approach to large eddy simulation of three-dimensional turbulence},
  author={Li, Zhijie and Peng, Wenhui and Yuan, Zelong and Wang, Jianchun},
  journal={Theoretical and Applied Mechanics Letters},
  volume={12},
  number={6},
  pages={100389},
  year={2022},
  publisher={Elsevier}
}

@inproceedings{liu2023nuno,
  title={Nuno: A general framework for learning parametric pdes with non-uniform data},
  author={Liu, Songming and Hao, Zhongkai and Ying, Chengyang and Su, Hang and Cheng, Ze and Zhu, Jun},
  booktitle={International Conference on Machine Learning},
  pages={21658--21671},
  year={2023},
  organization={PMLR}
}

@article{li2023geometry,
  title={Geometry-informed neural operator for large-scale 3d pdes},
  author={Li, Zongyi and Kovachki, Nikola and Choy, Chris and Li, Boyi and Kossaifi, Jean and Otta, Shourya and Nabian, Mohammad Amin and Stadler, Maximilian and Hundt, Christian and Azizzadenesheli, Kamyar and others},
  journal={Advances in Neural Information Processing Systems},
  volume={36},
  pages={35836--35854},
  year={2023}
}

@article{raissi2019physics,
  title={Physics-informed neural networks: A deep learning framework for solving forward and inverse problems involving nonlinear partial differential equations},
  author={Raissi, Maziar and Perdikaris, Paris and Karniadakis, George E},
  journal={Journal of Computational physics},
  volume={378},
  pages={686--707},
  year={2019},
  publisher={Elsevier}
}

@article{li2024physics,
  title={Physics-informed neural operator for learning partial differential equations},
  author={Li, Zongyi and Zheng, Hongkai and Kovachki, Nikola and Jin, David and Chen, Haoxuan and Liu, Burigede and Azizzadenesheli, Kamyar and Anandkumar, Anima},
  journal={ACM/IMS Journal of Data Science},
  volume={1},
  number={3},
  pages={1--27},
  year={2024},
  publisher={ACM New York, NY}
}

@article{wang2021learning,
  title={Learning the solution operator of parametric partial differential equations with physics-informed DeepONets},
  author={Wang, Sifan and Wang, Hanwen and Perdikaris, Paris},
  journal={Science advances},
  volume={7},
  number={40},
  pages={eabi8605},
  year={2021},
  publisher={American Association for the Advancement of Science}
}

@article{cuomo2022scientific,
  title={Scientific machine learning through physics--informed neural networks: Where we are and what’s next},
  author={Cuomo, Salvatore and Di Cola, Vincenzo Schiano and Giampaolo, Fabio and Rozza, Gianluigi and Raissi, Maziar and Piccialli, Francesco},
  journal={Journal of Scientific Computing},
  volume={92},
  number={3},
  pages={88},
  year={2022},
  publisher={Springer}
}

@article{cao2021choose,
  title={Choose a transformer: Fourier or galerkin},
  author={Cao, Shuhao},
  journal={Advances in neural information processing systems},
  volume={34},
  pages={24924--24940},
  year={2021}
}

@article{li2022transformer,
  title={Transformer for partial differential equations' operator learning},
  author={Li, Zijie and Meidani, Kazem and Farimani, Amir Barati},
  journal={arXiv preprint arXiv:2205.13671},
  year={2022}
}

@inproceedings{hao2023gnot,
  title={Gnot: A general neural operator transformer for operator learning},
  author={Hao, Zhongkai and Wang, Zhengyi and Su, Hang and Ying, Chengyang and Dong, Yinpeng and Liu, Songming and Cheng, Ze and Song, Jian and Zhu, Jun},
  booktitle={International Conference on Machine Learning},
  pages={12556--12569},
  year={2023},
  organization={PMLR}
}

@article{guibas2021adaptive,
  title={Adaptive fourier neural operators: Efficient token mixers for transformers},
  author={Guibas, John and Mardani, Morteza and Li, Zongyi and Tao, Andrew and Anandkumar, Anima and Catanzaro, Bryan},
  journal={arXiv preprint arXiv:2111.13587},
  year={2021}
}

@article{dosovitskiy2020image,
  title={An image is worth 16x16 words: Transformers for image recognition at scale},
  author={Dosovitskiy, Alexey},
  journal={arXiv preprint arXiv:2010.11929},
  year={2020}
}

@inproceedings{liu2021swin,
  title={Swin transformer: Hierarchical vision transformer using shifted windows},
  author={Liu, Ze and Lin, Yutong and Cao, Yue and Hu, Han and Wei, Yixuan and Zhang, Zheng and Lin, Stephen and Guo, Baining},
  booktitle={Proceedings of the IEEE/CVF international conference on computer vision},
  pages={10012--10022},
  year={2021}
}

@article{brandstetter2022message,
  title={Message passing neural PDE solvers},
  author={Brandstetter, Johannes and Worrall, Daniel and Welling, Max},
  journal={arXiv preprint arXiv:2202.03376},
  year={2022}
}

@inproceedings{ronneberger2015u,
  title={U-net: Convolutional networks for biomedical image segmentation},
  author={Ronneberger, Olaf and Fischer, Philipp and Brox, Thomas},
  booktitle={International Conference on Medical image computing and computer-assisted intervention},
  pages={234--241},
  year={2015},
  organization={Springer}
}

@article{radford2019language,
  title={Language models are unsupervised multitask learners},
  author={Radford, Alec and Wu, Jeffrey and Child, Rewon and Luan, David and Amodei, Dario and Sutskever, Ilya and others},
  journal={OpenAI blog},
  volume={1},
  number={8},
  pages={9},
  year={2019}
}

@inproceedings{devlin2019bert,
  title={Bert: Pre-training of deep bidirectional transformers for language understanding},
  author={Devlin, Jacob and Chang, Ming-Wei and Lee, Kenton and Toutanova, Kristina},
  booktitle={Proceedings of the 2019 conference of the North American chapter of the association for computational linguistics: human language technologies, volume 1 (long and short papers)},
  pages={4171--4186},
  year={2019}
}

@inproceedings{he2020momentum,
  title={Momentum contrast for unsupervised visual representation learning},
  author={He, Kaiming and Fan, Haoqi and Wu, Yuxin and Xie, Saining and Girshick, Ross},
  booktitle={Proceedings of the IEEE/CVF conference on computer vision and pattern recognition},
  pages={9729--9738},
  year={2020}
}

@inproceedings{he2022masked,
  title={Masked autoencoders are scalable vision learners},
  author={He, Kaiming and Chen, Xinlei and Xie, Saining and Li, Yanghao and Doll{\'a}r, Piotr and Girshick, Ross},
  booktitle={Proceedings of the IEEE/CVF conference on computer vision and pattern recognition},
  pages={16000--16009},
  year={2022}
}

@article{jumper2021highly,
  title={Highly accurate protein structure prediction with AlphaFold},
  author={Jumper, John and Evans, Richard and Pritzel, Alexander and Green, Tim and Figurnov, Michael and Ronneberger, Olaf and Tunyasuvunakool, Kathryn and Bates, Russ and {\v{Z}}{\'\i}dek, Augustin and Potapenko, Anna and others},
  journal={nature},
  volume={596},
  number={7873},
  pages={583--589},
  year={2021},
  publisher={Nature Publishing Group UK London}
}

@article{nguyen2023climax,
  title={Climax: A foundation model for weather and climate},
  author={Nguyen, Tung and Brandstetter, Johannes and Kapoor, Ashish and Gupta, Jayesh K and Grover, Aditya},
  journal={arXiv preprint arXiv:2301.10343},
  year={2023}
}

@inproceedings{zhou2023uni,
  title={Uni-mol: A universal 3d molecular representation learning framework},
  author={Zhou, Gengmo and Gao, Zhifeng and Ding, Qiankun and Zheng, Hang and Xu, Hongteng and Wei, Zhewei and Zhang, Linfeng and Ke, Guolin},
  booktitle={The eleventh international conference on learning representations},
  year={2023}
}

@article{mialon2023self,
  title={Self-supervised learning with lie symmetries for partial differential equations},
  author={Mialon, Gr{\'e}goire and Garrido, Quentin and Lawrence, Hannah and Rehman, Danyal and LeCun, Yann and Kiani, Bobak},
  journal={Advances in Neural Information Processing Systems},
  volume={36},
  pages={28973--29004},
  year={2023}
}

@article{subramanian2023towards,
  title={Towards foundation models for scientific machine learning: Characterizing scaling and transfer behavior},
  author={Subramanian, Shashank and Harrington, Peter and Keutzer, Kurt and Bhimji, Wahid and Morozov, Dmitriy and Mahoney, Michael W and Gholami, Amir},
  journal={Advances in Neural Information Processing Systems},
  volume={36},
  pages={71242--71262},
  year={2023}
}

@article{yang2023context,
  title={In-context operator learning with data prompts for differential equation problems},
  author={Yang, Liu and Liu, Siting and Meng, Tingwei and Osher, Stanley J},
  journal={Proceedings of the National Academy of Sciences},
  volume={120},
  number={39},
  pages={e2310142120},
  year={2023},
  publisher={National Academy of Sciences}
}

@article{mccabe2023multiple,
  title={Multiple physics pretraining for physical surrogate models},
  author={McCabe, Michael and Blancard, Bruno R{\'e}galdo-Saint and Parker, Liam Holden and Ohana, Ruben and Cranmer, Miles and Bietti, Alberto and Eickenberg, Michael and Golkar, Siavash and Krawezik, Geraud and Lanusse, Francois and others},
  journal={arXiv preprint arXiv:2310.02994},
  year={2023}
}

@article{hao2024dpot,
  title={Dpot: Auto-regressive denoising operator transformer for large-scale pde pre-training},
  author={Hao, Zhongkai and Su, Chang and Liu, Songming and Berner, Julius and Ying, Chengyang and Su, Hang and Anandkumar, Anima and Song, Jian and Zhu, Jun},
  journal={arXiv preprint arXiv:2403.03542},
  year={2024}
}

@article{chen2024omniarch,
  title={OmniArch: Building Foundation Model For Scientific Computing},
  author={Chen, Tianyu and Zhou, Haoyi and Li, Ying and Wang, Hao and Gao, Chonghan and Shi, Rongye and Zhang, Shanghang and Li, Jianxin},
  journal={arXiv preprint arXiv:2402.16014},
  year={2024}
}

@article{takamoto2022pdebench,
  title={Pdebench: An extensive benchmark for scientific machine learning},
  author={Takamoto, Makoto and Praditia, Timothy and Leiteritz, Raphael and MacKinlay, Daniel and Alesiani, Francesco and Pfl{\"u}ger, Dirk and Niepert, Mathias},
  journal={Advances in Neural Information Processing Systems},
  volume={35},
  pages={1596--1611},
  year={2022}
}

@article{gupta2022towards,
  title={Towards Multi-spatiotemporal-scale Generalized PDE Modeling},
  author={Gupta, Jayesh K and Brandstetter, Johannes},
  journal={arXiv preprint arXiv:2209.15616},
  year={2022}
}

@article{luo2023cfdbench,
  title={CFDBench: A Large-Scale Benchmark for Machine Learning Methods in Fluid Dynamics},
  author={Luo, Yining and Chen, Yingfa and Zhang, Zhen},
  journal={arXiv preprint arXiv:2310.05963},
  year={2023}
}

@inproceedings{sohl2015deep,
  title={Deep unsupervised learning using nonequilibrium thermodynamics},
  author={Sohl-Dickstein, Jascha and Weiss, Eric and Maheswaranathan, Niru and Ganguli, Surya},
  booktitle={International conference on machine learning},
  pages={2256--2265},
  year={2015},
  organization={pmlr}
}

@article{ho2020denoising,
  title={Denoising diffusion probabilistic models},
  author={Ho, Jonathan and Jain, Ajay and Abbeel, Pieter},
  journal={Advances in neural information processing systems},
  volume={33},
  pages={6840--6851},
  year={2020}
}

@article{song2020score,
  title={Score-based generative modeling through stochastic differential equations},
  author={Song, Yang and Sohl-Dickstein, Jascha and Kingma, Diederik P and Kumar, Abhishek and Ermon, Stefano and Poole, Ben},
  journal={arXiv preprint arXiv:2011.13456},
  year={2020}
}

@article{ho2022video,
  title={Video diffusion models},
  author={Ho, Jonathan and Salimans, Tim and Gritsenko, Alexey and Chan, William and Norouzi, Mohammad and Fleet, David J},
  journal={Advances in neural information processing systems},
  volume={35},
  pages={8633--8646},
  year={2022}
}

@article{karras2022elucidating,
  title={Elucidating the design space of diffusion-based generative models},
  author={Karras, Tero and Aittala, Miika and Aila, Timo and Laine, Samuli},
  journal={Advances in neural information processing systems},
  volume={35},
  pages={26565--26577},
  year={2022}
}

@article{chen2018neural,
  title={Neural ordinary differential equations},
  author={Chen, Ricky TQ and Rubanova, Yulia and Bettencourt, Jesse and Duvenaud, David K},
  journal={Advances in neural information processing systems},
  volume={31},
  year={2018}
}

@article{liu2022flow,
  title={Flow straight and fast: Learning to generate and transfer data with rectified flow},
  author={Liu, Xingchao and Gong, Chengyue and Liu, Qiang},
  journal={arXiv preprint arXiv:2209.03003},
  year={2022}
}

@article{jia2025cod,
  title={CoD: A Diffusion Foundation Model for Image Compression},
  author={Jia, Zhaoyang and Zheng, Zihan and Xue, Naifu and Li, Jiahao and Li, Bin and Guo, Zongyu and Zhang, Xiaoyi and Li, Houqiang and Lu, Yan},
  journal={arXiv preprint arXiv:2511.18706},
  year={2025}
}

@article{su2024roformer,
  title={Roformer: Enhanced transformer with rotary position embedding},
  author={Su, Jianlin and Ahmed, Murtadha and Lu, Yu and Pan, Shengfeng and Bo, Wen and Liu, Yunfeng},
  journal={Neurocomputing},
  volume={568},
  pages={127063},
  year={2024},
  publisher={Elsevier}
}

@article{han2018solving,
  title={Solving high-dimensional partial differential equations using deep learning},
  author={Han, Jiequn and Jentzen, Arnulf and E, Weinan},
  journal={Proceedings of the National Academy of Sciences},
  volume={115},
  number={34},
  pages={8505--8510},
  year={2018},
  publisher={National Academy of Sciences}
}

@article{ye2024pdeformer,
  title={Pdeformer: Towards a foundation model for one-dimensional partial differential equations},
  author={Ye, Zhanhong and Huang, Xiang and Chen, Leheng and Liu, Hongsheng and Wang, Zidong and Dong, Bin},
  journal={arXiv preprint arXiv:2402.12652},
  year={2024}
}

@article{wang2024beno,
  title={Beno: Boundary-embedded neural operators for elliptic pdes},
  author={Wang, Haixin and Li, Jiaxin and Dwivedi, Anubhav and Hara, Kentaro and Wu, Tailin},
  journal={arXiv preprint arXiv:2401.09323},
  year={2024}
}

@article{wan2025spectral,
  title={Spectral-inspired Neural Operator for Data-efficient PDE Simulation in Physics-agnostic Regimes},
  author={Wan, Han and Zhang, Rui and Sun, Hao},
  journal={arXiv preprint arXiv:2505.21573},
  year={2025}
}

@article{zhang2025omnifluids,
  title={OmniFluids: Unified Physics Pre-trained Modeling of Fluid Dynamics},
  author={Zhang, Rui and Meng, Qi and Wan, Han and Liu, Yang and Ma, Zhi-Ming and Sun, Hao},
  journal={arXiv preprint arXiv:2506.10862},
  year={2025}
}

@inproceedings{fengheap,
  title={HEAP: Hyper Extended A-PDHG Operator for Constrained High-dim PDEs},
  author={Feng, Mingquan and Liao, Weixin and Huang, Yixin and Fu, Yifan and Zheng, Qifu and Yan, Junchi},
  booktitle={Forty-second International Conference on Machine Learning}
}

@inproceedings{rahaman2019spectral,
  title={On the spectral bias of neural networks},
  author={Rahaman, Nasim and Baratin, Aristide and Arpit, Devansh and Draxler, Felix and Lin, Min and Hamprecht, Fred and Bengio, Yoshua and Courville, Aaron},
  booktitle={International conference on machine learning},
  pages={5301--5310},
  year={2019},
  organization={PMLR}
}

@article{khodakarami2025mitigating,
  title={Mitigating spectral bias in neural operators via high-frequency scaling for physical systems},
  author={Khodakarami, Siavash and Oommen, Vivek and Bora, Aniruddha and Karniadakis, George Em},
  journal={arXiv preprint arXiv:2503.13695},
  year={2025}
}

@article{lipman2022flow,
  title={Flow matching for generative modeling},
  author={Lipman, Yaron and Chen, Ricky TQ and Ben-Hamu, Heli and Nickel, Maximilian and Le, Matt},
  journal={arXiv preprint arXiv:2210.02747},
  year={2022}
}

@article{herde2024poseidon,
  title={Poseidon: Efficient foundation models for pdes},
  author={Herde, Maximilian and Raonic, Bogdan and Rohner, Tobias and K{\"a}ppeli, Roger and Molinaro, Roberto and de B{\'e}zenac, Emmanuel and Mishra, Siddhartha},
  journal={Advances in Neural Information Processing Systems},
  volume={37},
  pages={72525--72624},
  year={2024}
}

@article{price2023gencast,
  title={Gencast: Diffusion-based ensemble forecasting for medium-range weather},
  author={Price, Ilan and Sanchez-Gonzalez, Alvaro and Alet, Ferran and Andersson, Tom R and El-Kadi, Andrew and Masters, Dominic and Ewalds, Timo and Stott, Jacklynn and Mohamed, Shakir and Battaglia, Peter and others},
  journal={arXiv preprint arXiv:2312.15796},
  year={2023}
}

@inproceedings{davtyan2023efficient,
  title={Efficient video prediction via sparsely conditioned flow matching},
  author={Davtyan, Aram and Sameni, Sepehr and Favaro, Paolo},
  booktitle={Proceedings of the IEEE/CVF International Conference on Computer Vision},
  pages={23263--23274},
  year={2023}
}

@article{ruhling2023dyffusion,
  title={Dyffusion: A dynamics-informed diffusion model for spatiotemporal forecasting},
  author={R{\"u}hling Cachay, Salva and Zhao, Bo and Joren, Hailey and Yu, Rose},
  journal={Advances in neural information processing systems},
  volume={36},
  pages={45259--45287},
  year={2023}
}

@article{hu2024wavelet,
  title={Wavelet diffusion neural operator},
  author={Hu, Peiyan and Wang, Rui and Zheng, Xiang and Zhang, Tao and Feng, Haodong and Feng, Ruiqi and Wei, Long and Wang, Yue and Ma, Zhi-Ming and Wu, Tailin},
  journal={arXiv preprint arXiv:2412.04833},
  year={2024}
}

@article{yang2023denoising,
  title={A denoising diffusion model for fluid field prediction},
  author={Yang, Gefan and Sommer, Stefan},
  journal={arXiv preprint arXiv:2301.11661},
  year={2023}
}

@inproceedings{peebles2023scalable,
  title={Scalable diffusion models with transformers},
  author={Peebles, William and Xie, Saining},
  booktitle={Proceedings of the IEEE/CVF international conference on computer vision},
  pages={4195--4205},
  year={2023}
}

@article{he2022latent,
  title={Latent video diffusion models for high-fidelity long video generation},
  author={He, Yingqing and Yang, Tianyu and Zhang, Yong and Shan, Ying and Chen, Qifeng},
  journal={arXiv preprint arXiv:2211.13221},
  year={2022}
}

@article{ho2022classifier,
  title={Classifier-free diffusion guidance},
  author={Ho, Jonathan and Salimans, Tim},
  journal={arXiv preprint arXiv:2207.12598},
  year={2022}
}

@article{lu2024generative,
  title={Generative downscaling of PDE solvers with physics-guided diffusion models},
  author={Lu, Yulong and Xu, Wuzhe},
  journal={Journal of scientific computing},
  volume={101},
  number={3},
  pages={71},
  year={2024},
  publisher={Springer}
}

@article{zhang2025physics,
  title={Physics-Informed Distillation of Diffusion Models for PDE-Constrained Generation},
  author={Zhang, Yi and Zou, Difan},
  journal={arXiv preprint arXiv:2505.22391},
  year={2025}
}

@article{huang2024diffusionpde,
  title={DiffusionPDE: Generative PDE-solving under partial observation},
  author={Huang, Jiahe and Yang, Guandao and Wang, Zichen and Park, Jeong Joon},
  journal={Advances in Neural Information Processing Systems},
  volume={37},
  pages={130291--130323},
  year={2024}
}

@article{jiang2025sig,
  title={Sig-DEG for Distillation: Making Diffusion Models Faster and Lighter},
  author={Jiang, Lei and Ge, Wen and Cariou-Kotlarek, Niels and Yi, Mingxuan and Chen, Po-Yu and Yang, Lingyi and Buet-Golfouse, Francois and Mittal, Gaurav and Ni, Hao},
  journal={arXiv preprint arXiv:2508.16939},
  year={2025}
}
\bibliographystyle{icml2026}

\newpage
\appendix
\onecolumn
\section{Dataset Details}
\label{app:datasets}

\subsection{PDEBench Overview and Data Format}
We build our pre-training and evaluation subsets primarily on PDEBench.
Each dataset file is stored as a multi-array tensor with shape $(N, T, X, Y, Z, V)$, where
$N$ is the number of trajectories, $T$ the number of time steps, $(X,Y,Z)$ the spatial grid resolution,
and $V$ the number of physical variables (treated as channels).
Following the common PDEBench protocol, we split trajectories into 90\% training and 10\% test sets.

\subsection{PDE Families and Quantities of Interest}
Below we summarize the PDEs used in this work and the associated physical variables.
We denote the velocity by $\mathbf{v}$ (or $u$ in 1D), density by $\rho$, pressure by $p$,
and water depth by $h$.

\paragraph{1D Linear Advection (1D\_Advection).}
We consider the 1D advection equation
\begin{equation}
\partial_t u + \beta \, \partial_x u = 0,
\end{equation}
where $u(x,t)$ is the transported scalar/velocity and $\beta$ is the advection speed.

\paragraph{1D Viscous Burgers (1D\_Burgers).}
We consider the viscous Burgers equation
\begin{equation}
\partial_t u + u\,\partial_x u = \nu\,\partial_{xx}u,
\end{equation}
where $\nu$ is the viscosity.

\paragraph{Compressible Navier--Stokes (CFD in 1D/2D/3D).}
For compressible flows, PDEBench uses an isothermal compressible Navier--Stokes system
\begin{align}
\partial_t \rho + \nabla\cdot(\rho \mathbf{v}) &= 0, \\
\rho\big(\partial_t \mathbf{v} + \mathbf{v}\cdot\nabla \mathbf{v}\big)
&= -\nabla p + \eta\,\Delta \mathbf{v} + \Big(\zeta + \frac{\eta}{3}\Big)\nabla(\nabla\cdot \mathbf{v}),
\end{align}
where $\eta$ and $\zeta$ are viscosity coefficients.
We use $(\rho, p, \mathbf{v})$ as the predicted physical quantities for all 1D/2D/3D CFD subsets.

\paragraph{2D Reaction--Diffusion (diff\_react\_2d).}
We use the 2D activator--inhibitor system (FitzHugh--Nagumo reactions) with variables $(u,v)$,
\begin{align}
\partial_t u &= D_u \Delta u + R_u(u,v), \\
\partial_t v &= D_v \Delta v + R_v(u,v),
\end{align}
where $R_u, R_v$ are the reaction terms and $D_u,D_v$ the diffusion coefficients.

\paragraph{2D Shallow-Water Equations (swe\_pdb).}
We use the 2D SWE system
\begin{align}
\partial_t h + \nabla\cdot(h\mathbf{u}) &= 0, \\
\partial_t(h\mathbf{u}) + \nabla\cdot\!\Big(h\mathbf{u}\mathbf{u}^\top + \tfrac{1}{2}g_r h^2 \mathbf{I}\Big)
&= -g_r h \nabla b,
\end{align}
where $h$ is water depth, $\mathbf{u}$ is depth-averaged velocity, $b$ is bathymetry, and $g_r$ is gravity.

\paragraph{2D Incompressible Navier--Stokes (ns2d\_incom).}
We use the incompressible Navier--Stokes equations with an external forcing term $\mathbf{f}$,
\begin{align}
\nabla\cdot \mathbf{v} &= 0, \\
\partial_t \mathbf{v} + \mathbf{v}\cdot\nabla \mathbf{v} &= -\nabla p + \nu \Delta \mathbf{v} + \mathbf{f}.
\end{align}

\subsection{Subsets Used in This Work and Pre-processing}
\label{app:datasets:subsets}

\paragraph{Subset selection.}
In our implementation, we use the following PDEBench subsets:
\texttt{1D\_Advection}, \texttt{1D\_Burgers}, \texttt{1D\_CFD};
\texttt{ns2d\_pdb\_M1\_eta1e-1\_zeta1e-1}, \texttt{ns2d\_pdb\_M1\_eta1e-2\_zeta1e-2},
\texttt{ns2d\_pdb\_M1e-1\_eta1e-1\_zeta1e-1}, \texttt{ns2d\_pdb\_M1e-1\_eta1e-2\_zeta1e-2},
\texttt{swe\_pdb}, \texttt{diff\_react\_2d}, \texttt{ns2d\_incom};
and \texttt{ns3d\_pdb\_M1\_turb}, \texttt{ns3d\_pdb\_M1\_rand}.

\paragraph{Unified target resolution.}
To support unified spatiotemporal tensor processing, we resample each trajectory to a fixed target resolution per dimension:
1D: $1024$ points; 2D: $128\times128$; 3D: $64\times64\times64$.

\begin{table*}[t]
\vspace{-1mm}
\centering
\small
\setlength{\tabcolsep}{6pt}
\renewcommand{\arraystretch}{1.08}
\caption{PDEBench subsets used in this work (variables and resolutions). $N$/$T$/$N_s$ denote \#trajectories, \#timesteps, and spatial grid resolution in the PDEBench release; \texttt{Target $N_s$} is our unified resolution.}
\label{tab:pdebench_used}
\begin{tabular}{l c c c c c}
\toprule
Subset & Dim & physical variables & $N$ & $(T, N_s)$ & Target $N_s$ \\
\midrule
\texttt{1D\_Advection} & 1D & $(V_x)$ & 80000 & $(201,\;1024)$ & $1024$ \\
\texttt{1D\_Burgers}   & 1D & $(V_x)$ & 120000 & $(201,\;1024)$ & $1024$ \\
\texttt{1D\_CFD} (CNS) & 1D & $(V_x,\rho,p)$ & 50000 & $(101,\;1024)$ & $1024$ \\
\midrule
\texttt{2D\_CFD} (CNS) & 2D & $(V_x,V_y,\rho,p)$ & 40000  & $(21,\;512^2)$ & $128^2$ \\
\texttt{diff\_react\_2d} & 2D & $(u,v)$ & 1000 & $(101,\;128^2)$ & $128^2$ \\
\texttt{swe\_pdb} & 2D & $(h)$ & 1000 & $(101,\;128^2)$ & $128^2$ \\
\texttt{ns2d\_incom} & 2D & $(V_x,V_y,\texttt{particles})$ & 1000 & $(1001,\;256^2)$ & $128^2$ \\
\midrule
\texttt{ns3d\_pdb} (CNS) & 3D & $(V_x,V_y,V_z,\rho,p)$ & 700 & $(21,\;128^3)$ & $64^3$ \\
\bottomrule
\end{tabular}
\vspace{-2mm}
\end{table*}

\subsection{Datasets for Zero-shot Generalization}
\label{app:datasets:zeroshot}

To assess out-of-domain generalization without any additional training, we evaluate our foundation model in a zero-shot manner on several \emph{unseen} PDEBench cases.
All zero-shot datasets considered here are 2D \emph{compressible} Navier--Stokes flows, but correspond to distinct physical phenomena and regimes that are not included in our pre-training suite.

\paragraph{2D Shock tube (Shock).}
This dataset features shock waves characterized by abrupt discontinuities in flow properties (e.g., density and pressure) and strong nonlinearity.
We predict the standard compressible-flow quantities $(\rho, p, v_x, v_y)$.
The data are provided as a high-resolution spatiotemporal trajectory (e.g., $1024\times1024$ over $\sim$100 timesteps in the PDEBench release).

\paragraph{2D Kelvin--Helmholtz Instability (KH).}
The KH dataset describes interfacial instability arising from velocity shear (and/or density contrast) between two layers, leading to vortex roll-up and mixing.
We again predict $(\rho, p, v_x, v_y)$.
The dataset contains multiple trajectories generated with different physical parameters (e.g., Mach number and Reynolds-related coefficients) and is evaluated by averaging across all provided samples.

\paragraph{Orszag--Tang Vortex (OTVortex).}
The Orszag--Tang vortex system is a canonical benchmark for compressible flows with complex interacting vortical structures.
It exhibits rapid formation of shocks and vortices from carefully designed initial conditions.
We predict $(\rho, p, v_x, v_y)$ on the full trajectory (high-resolution spatial grids and $\sim$100 timesteps in the PDEBench release).

\section{Training Details}
\label{sec:training_details}

\paragraph{Optimization.}
Unless otherwise specified, we use a single set of optimization hyperparameters across all datasets and model sizes.
We train end-to-end with AdamW using a base learning rate of $10^{-4}$ and weight decay $10^{-4}$ for 200 epochs.
The learning rate follows linear warmup for 5\% of the training steps and then cosine decay to a minimum learning rate of $10^{-6}$.
We enable mixed-precision training and apply gradient clipping with a global norm of 1.0 for stability.
We validate every 5 epochs and periodically checkpoint the model during training.

\paragraph{Noise level sampling and guidance.}
Following the conditional flow-matching formulation, we sample the continuous noise level $t$ from a logit-normal distribution
(parameterized by $P_{\text{mean}}=-0.8$, $P_{\text{std}}=0.8$) within $t\in[T_{\min},T_{\max}]$ where $T_{\min}=10^{-4}$ and $T_{\max}=1$.
We apply condition dropout with probability 0.1 to enable classifier-free guidance (CFG).
At inference, we solve the corresponding probability-flow ODE using a fixed-step solver (Euler by default; Heun when specified),
with 40 steps for final sampling (and 10 steps for fast evaluation).
We use $\mathrm{CFG}$ scale 2.0 in our default setting.
In addition, we sample $t=0$ with probability 0.1 during training to explicitly encourage reconstruction.

\paragraph{Data preprocessing.}
We normalize each channel using statistics computed from the conditioning history window,
and keep a unified variable vocabulary with a binary channel mask.
To support unified training across dimensions, all trajectories are resampled to a fixed target resolution per dimensionality:
1024 points for 1D, $128^2$ for 2D, and $64^3$ for 3D.
We use a fixed history length of 10 frames and predict a horizon of 10 frames  following previous works.

\paragraph{Batching and sampling across dimensions.}
To balance compute and memory across heterogeneous dimensionalities, we use per-dimension batch sizes.
Across all model sizes (S/M/L/XL), we use batch sizes $\{16,8,4\}$ for \{1D, 2D, 3D\} respectively.
When pre-training on the mixed suite, we sample 1D and 2D subsets with uniform weights, and upweight 3D subsets by a factor of 5 to compensate for their smaller corpus size.

\paragraph{Model scaling.}
We train four model sizes (S/M/L/XL) with the same data mixture and training recipe, and scale capacity by increasing the embedding width as well as the depth of both the encoder and the flow-matching operator.
All variants share the same 4D patch size $(p_t,p_h,p_w,p_d)=(2,8,8,8)$, in the encoder, and the MLP ratio of 4.0.
The attention head numbers in both modules follow the model width (see Table~\ref{tab:model_sizes}).

\begin{table}[t]
\vspace{-1mm}
\centering
\small
\setlength{\tabcolsep}{3.8pt}
\renewcommand{\arraystretch}{1.08}
\caption{Model scaling configurations used in this work.}
\label{tab:model_sizes}
\begin{tabular}{lccccccc}
\toprule
Model & $d$ & $L_e$ & $h_e$ & $L_j$ & $h_j$ & Params (M) &  Encoder/Flow-Matching Operator (M) \\
\midrule
S  & 256  & 4  & 4  & 7  & 4  & 22.77  & 6.11 / 16.66 \\
M  & 384  & 6  & 6  & 10 & 6  & 60.49  & 15.51 / 44.98 \\
L  & 512  & 8  & 8  & 14 & 8  & 135.31 & 32.29 / 103.01 \\
XL & 1024 & 8  & 8  & 14 & 8  & 512.28 & 119.64 / 392.65 \\
\bottomrule
\end{tabular}
\vspace{-2mm}
\end{table}

\paragraph{Unified training with $t{=}0$ samples.}
Our model is trained with a rectified-flow/flow-matching velocity loss, but is parameterized to predict the clean field.
We adopt the \emph{unified training} strategy of CoD\cite{jia2025cod} to explicitly inject a distortion-oriented term by occasionally sampling the fully noised endpoint $t{=}0$.
Concretely, we use the linear interpolation path
\begin{equation}
\label{eq:rf_path}
z_t \;=\; t\,x + (1-t)\,\epsilon,\qquad t\in[0,1],\;\epsilon\sim\mathcal{N}(0,I),
\end{equation}
where the rectified-flow velocity is $v = x-\epsilon$.
Our network outputs an $x$-prediction $\hat{x}_\theta=f_\theta(z_t,t;\,c)$, which is analytically converted to the velocity used by the sampler:
\begin{equation}
\label{eq:xpred_to_vpred}
\hat v_\theta(z_t,t;\,c)
\;=\;
\frac{\hat x_\theta(z_t,t;\,c)-z_t}{1-t}.
\end{equation}
We then minimize the standard velocity regression objective
\begin{equation}
\label{eq:v_loss}
\mathcal{L}_{v}
\;=\;
\mathbb{E}_{x,\epsilon,t}\Big[\big\|\hat v_\theta(z_t,t;\,c) - (x-\epsilon)\big\|_2^2\Big].
\end{equation}

Following CoD\cite{jia2025cod}, with probability $p_0$ we set $t{=}0$ (in our implementation, $p_0{=}0.1$), and otherwise sample $t$ from the default continuous time distribution.
At $t{=}0$, since $z_0=\epsilon$, Eq.~\eqref{eq:xpred_to_vpred} becomes $\hat v_\theta=\hat x_\theta-\epsilon$, and the velocity loss reduces to a one-step reconstruction loss:
\begin{equation}
\label{eq:t0_recon}
\mathcal{L}_{v}\big|_{t=0}
\;=\;
\big\|(\hat x_\theta-\epsilon) - (x-\epsilon)\big\|_2^2
\;=\;
\|\hat x_\theta-x\|_2^2.
\end{equation}
This provides explicit distortion supervision (aligned with our nRMSE-style evaluation) while keeping the same ODE-based sampling at inference, since we still evolve $z_t$ by $\mathrm{d}z_t/\mathrm{d}t=\hat v_\theta(z_t,t;\,c)$.

\section{More Results}
\label{app:more_results}
\subsection{Multi-resolution support via unified representation and 4D RoPE}
\label{app:more_results:multires}

Our model operates on a unified 4D spatiotemporal representation (Sec.~\ref{sec:unified-interface}), where each patch token carries explicit integer coordinates $(t,h,w,d)$.
Because positional encoding is injected through coordinate-parameterized \emph{4D RoPE}, the same Transformer operator can be applied to inputs of different spatial resolutions (i.e., different token grids) without changing the architecture.
We evaluate a single trained model on 1D Burgers with different target resolutions; Table~\ref{tab:more_results:multires_1d_burgers} shows consistently low nRMSE across resolutions.

\begin{table}[t]
\centering
\small
\setlength{\tabcolsep}{10pt}
\renewcommand{\arraystretch}{1.1}
\caption{Multi-resolution evaluation on 1D Burgers (nRMSE $\downarrow$, 110 test samples).}
\label{tab:more_results:multires_1d_burgers}
\begin{tabular}{cc}
\toprule
Target resolution & nRMSE $\downarrow$ \\
\midrule
128  & 0.01077 \\
256  & 0.01063 \\
512  & 0.01082 \\
768  & 0.01099 \\
1024 & 0.01033 \\
\bottomrule
\end{tabular}
\vspace{-2mm}
\end{table}

\subsection{Inference time}
\label{app:more_results:inference_time}

We report wall-clock inference time for generating a \emph{10-frame future window in one shot} (i.e., a single sampling run produces 10 frames simultaneously, rather than autoregressive rollout).
We measure latency on \texttt{1D\_Advection} with \texttt{batch\_size=1} and \texttt{prediction\_steps=10}, varying the number of ODE solver steps.

\begin{table}[t]
\centering
\small
\setlength{\tabcolsep}{8pt}
\renewcommand{\arraystretch}{1.1}
\caption{Inference time (seconds) for one-shot generation of 10 frames on \texttt{1D\_Advection}. Mean $\pm$ std over 10 runs.}
\label{tab:more_results:inference_time_1d_adv}
\begin{tabular}{lccc}
\toprule
Model & 1 step & 5 steps & 40 steps \\
\midrule
UniFluids-S  & 0.03617 $\pm$ 0.00023 & 0.16240 $\pm$ 0.00060 & 1.26670 $\pm$ 0.00355 \\
UniFluids-M  & 0.05475 $\pm$ 0.00594 & 0.23163 $\pm$ 0.01024 & 1.80871 $\pm$ 0.01983 \\
UniFluids-XL & 0.06911 $\pm$ 0.00062 & 0.31227 $\pm$ 0.00126 & 2.45982 $\pm$ 0.03195 \\
\bottomrule
\end{tabular}
\vspace{-2mm}
\end{table}

\subsection{Visualizations.}
For each dataset, we visualize a representative rollout segment from $T{+}1$ to $T{+}6$.
The top row shows the ground-truth future fields, and the bottom row shows the corresponding predictions from our model
(visualizing one representative variable/channel from the available physical variables).


\begin{figure*}[t]
\vspace{-1mm}
\centering
\begin{minipage}[t]{0.32\textwidth}
  \centering
  \includegraphics[width=\linewidth]{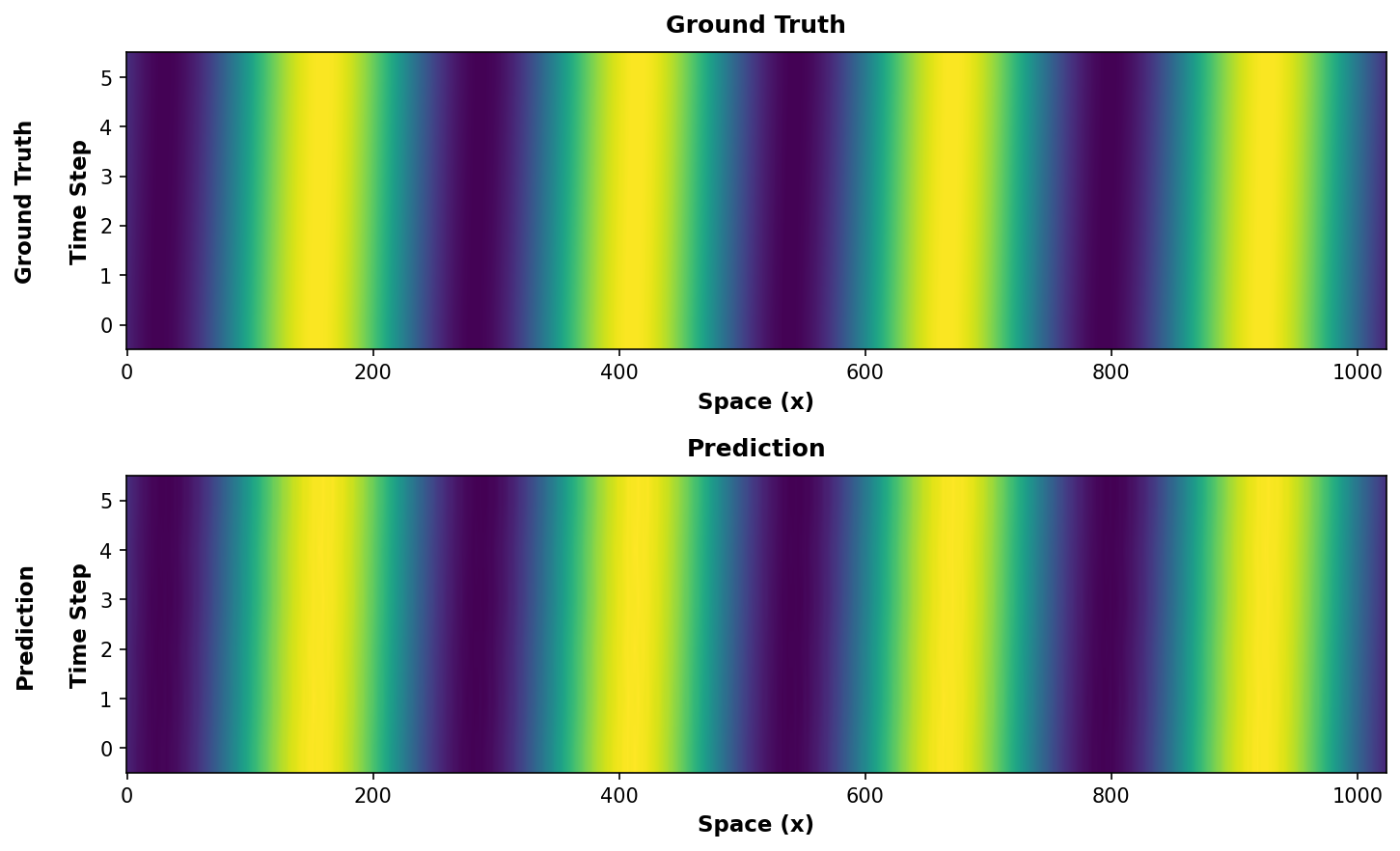}
\end{minipage}\hfill
\begin{minipage}[t]{0.32\textwidth}
  \centering
  \includegraphics[width=\linewidth]{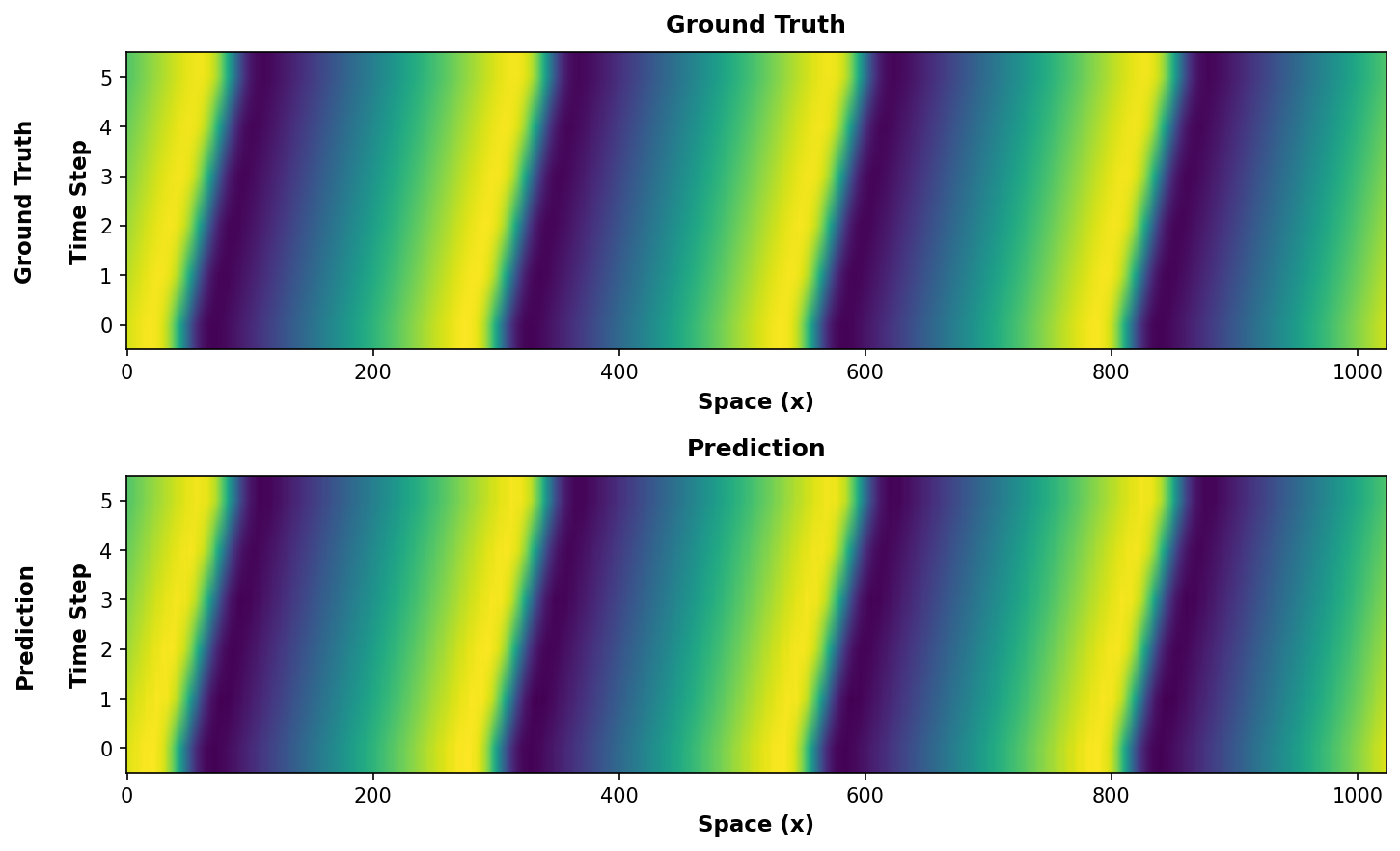}
\end{minipage}\hfill
\begin{minipage}[t]{0.32\textwidth}
  \centering
  \includegraphics[width=\linewidth]{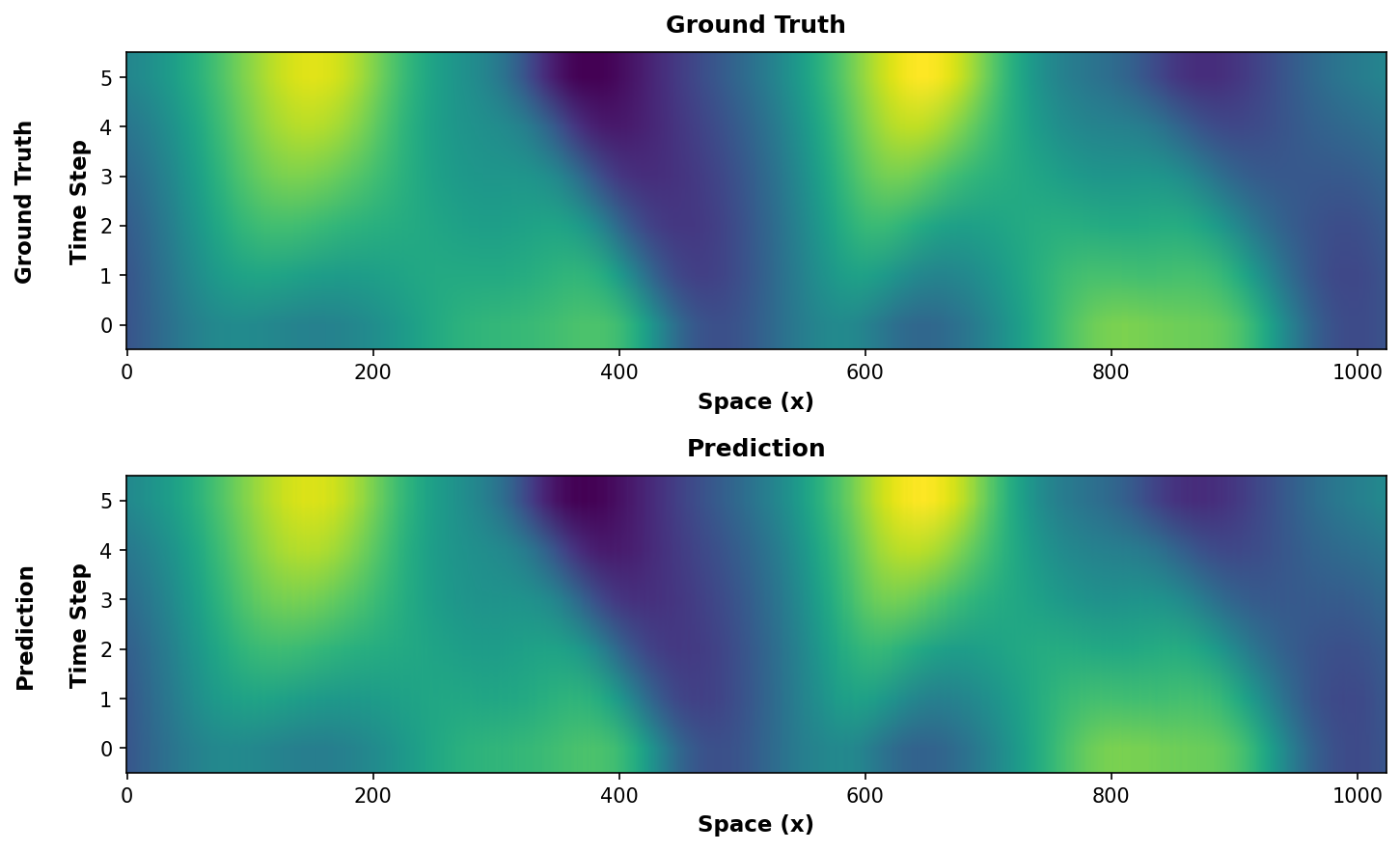}
\end{minipage}
\vspace{-2mm}
\caption{Prediction results on 1D PDEBench subsets.
From left to right: 1D Advection, 1D Burgers, and 1D CFD (compressible Navier--Stokes).
Each panel shows rollouts for time steps $T{+}1$ to $T{+}6$; the top row is ground truth and the bottom row is our prediction.}
\label{fig:more_qual_1d}
\vspace{-2mm}
\end{figure*}

\label{app:more_results:2d}

\begin{figure*}[t]
\vspace{-1mm}
\centering
\begin{minipage}[t]{0.49\textwidth}
  \centering
  \includegraphics[width=\linewidth]{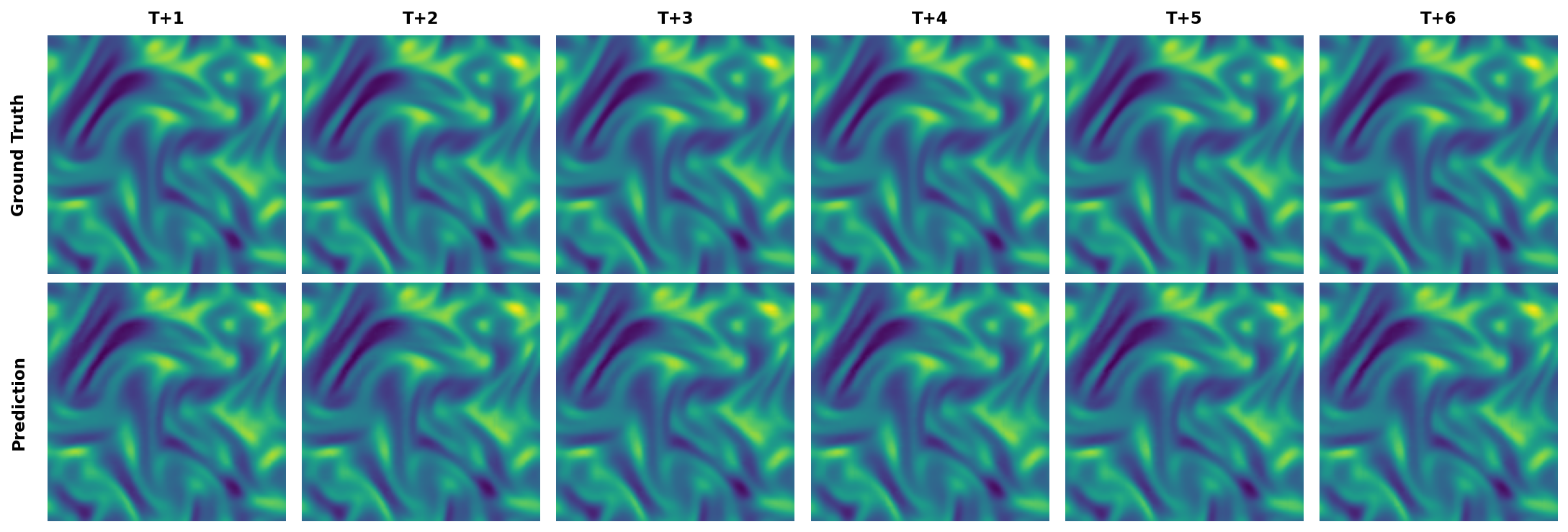}
\end{minipage}\hfill
\begin{minipage}[t]{0.49\textwidth}
  \centering
  \includegraphics[width=\linewidth]{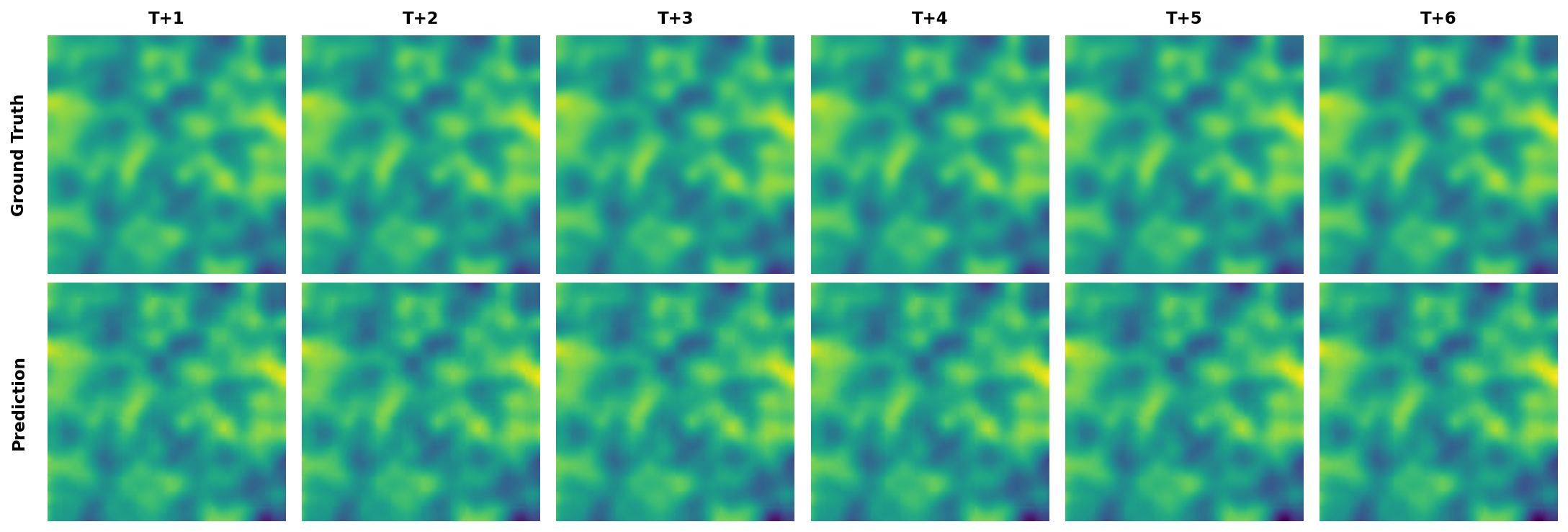}
\end{minipage}

\vspace{1mm}

\begin{minipage}[t]{0.49\textwidth}
  \centering
  \includegraphics[width=\linewidth]{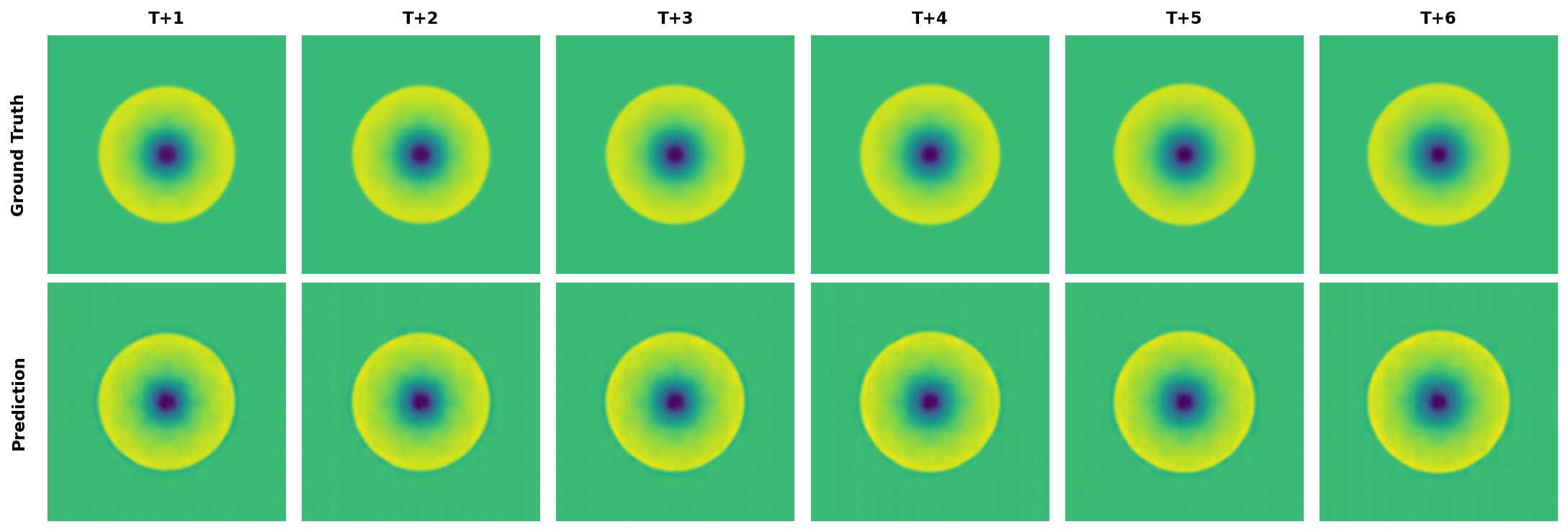}
\end{minipage}\hfill
\begin{minipage}[t]{0.49\textwidth}
  \centering
  \includegraphics[width=\linewidth]{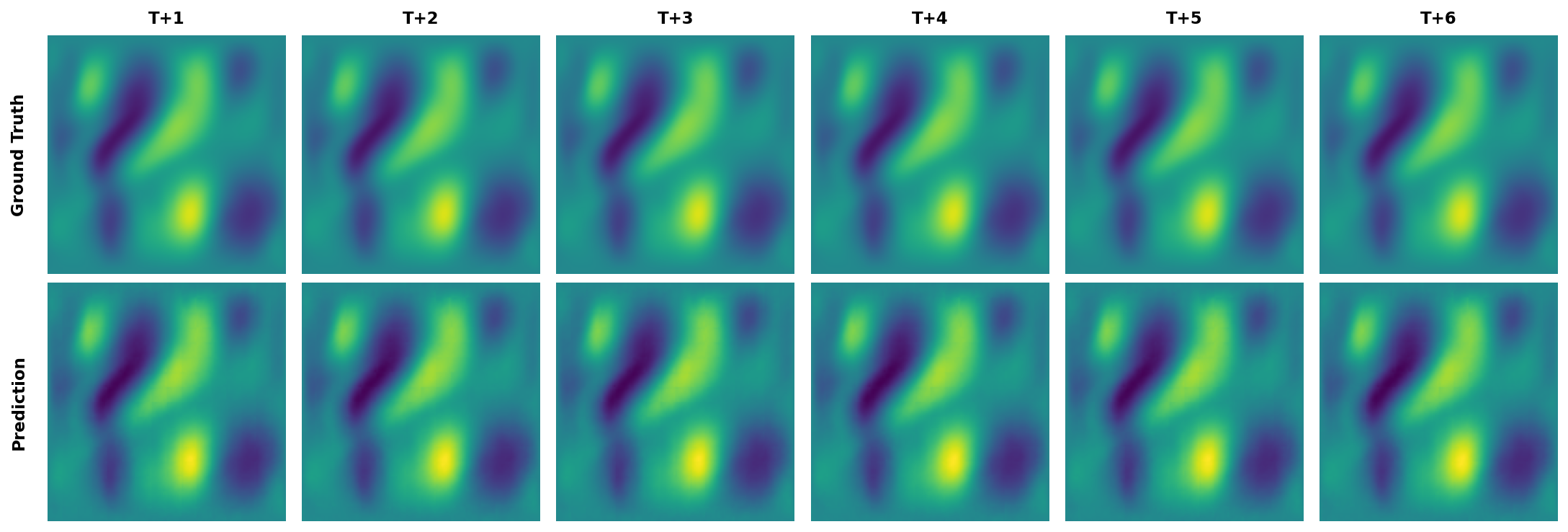}
\end{minipage}
\vspace{-2mm}
\caption{Prediction results on 2D PDEBench subsets.
Top row: ground truth; bottom row: our prediction; shown for time steps $T{+}1$ to $T{+}6$.
We visualize (top-left) 2D CFD (compressible Navier--Stokes), (top-right) reaction--diffusion,
(bottom-left) shallow-water equation, and (bottom-right) 2D incompressible Navier--Stokes (with particle tracer).}
\label{fig:more_qual_2d}
\vspace{-2mm}
\end{figure*}

\label{app:more_results:3d}

\begin{figure}[t]
\vspace{-1mm}
\centering
\includegraphics[width=\linewidth]{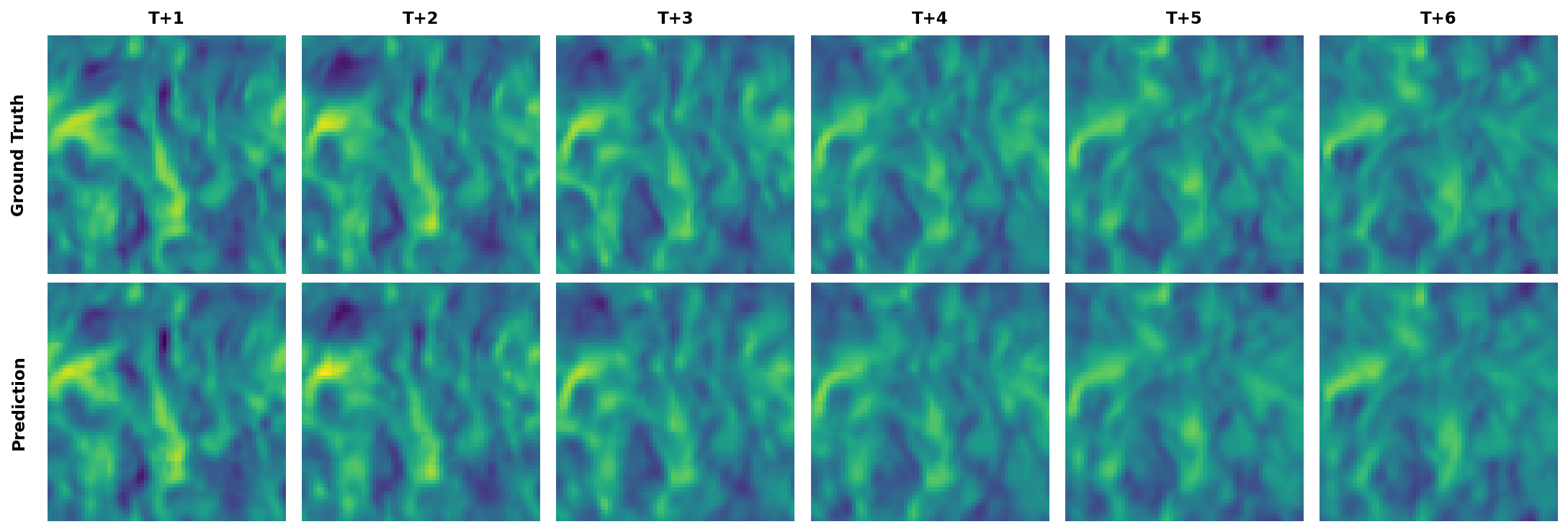}
\vspace{-2mm}
\caption{Prediction results on the 3D CFD subset (compressible Navier--Stokes).
We display rollouts for time steps $T{+}1$ to $T{+}6$; the top row shows ground truth and the bottom row shows our prediction.}
\label{fig:more_qual_3d}
\vspace{-2mm}
\end{figure}

\section{Limitations}
\label{app:limitations}

\paragraph{Resolution generalization is not fully free.}
While the unified 4D representation and 4D RoPE enable applying the same architecture across different token grids, performance can still degrade when the test resolution differs substantially from the training distribution, especially for sharp structures (e.g., shocks) and turbulence at high Reynolds numbers.

\paragraph{Sampling cost and deployment.}
Our flow-matching operator requires iterative ODE sampling; although this avoids autoregressive rollout and enables parallel generation of a future window, the wall-clock latency still scales with the number of solver steps.
Exploring stronger distillation/acceleration (e.g., fewer-step samplers) is left for future work.




\end{document}